\pdfoutput=1

\documentclass[11pt]{article}

\usepackage[table]{xcolor}
\usepackage{acl}

\usepackage{graphics}
\usepackage{booktabs}
\usepackage{times}
\usepackage{latexsym}
\usepackage[T1]{fontenc}
\usepackage[utf8]{inputenc}
\usepackage{microtype}
\usepackage{hyperref}
\usepackage{longtable}
\usepackage{multirow}
\usepackage{graphicx}
\usepackage{float}
\usepackage{cleveref}
\usepackage{balance}
\usepackage{enumitem}

\usepackage{tabularx,colortbl}

\usepackage{tikz}
\usepackage{collcell}

\usepackage{etoolbox}

\newtoggle{inTableHeader}
\toggletrue{inTableHeader}
\newcommand*{\StartTableHeader}{\global\toggletrue{inTableHeader}}%
\newcommand*{\EndTableHeader}{\global\togglefalse{inTableHeader}}%

\let\OldTabular\tabular%
\let\OldEndTabular\endtabular%
\renewenvironment{tabular}{\StartTableHeader\OldTabular}{\OldEndTabular\StartTableHeader}%

\newcommand*{\MinNumber}{0.0}%
\newcommand*{\MidNumber}{30.0} %
\newcommand*{\MaxNumber}{100.0}%

\newcommand{\ApplyGradient}[1]{%
  \iftoggle{inTableHeader}{#1}{
    \ifdim #1 pt > \MidNumber pt
        \pgfmathsetmacro{\PercentColor}{max(min(100.0*(#1 - \MidNumber)/(\MaxNumber-\MidNumber),100.0),0.00)} %
        \hspace{-0.33em}\colorbox{yellow!\PercentColor!blue}{#1}
    \else
        \pgfmathsetmacro{\PercentColor}{max(min(100.0*(\MidNumber - #1)/(\MidNumber-\MinNumber),100.0),0.00)} %
        \hspace{-0.33em}\colorbox{blue!\PercentColor!blue}{#1}
    \fi
  }}
\newcolumntype{R}{>{\collectcell\ApplyGradient}c<{\endcollectcell}}

\title{A Data-centric Framework for Improving Domain-specific \\ Machine Reading Comprehension Datasets}

\author{Iva Bojic$^{1}$ \and Josef Halim$^{1}$ \and Verena Suharman$^{1}$ \and Sreeja Tar$^{1}$ \and \\ {\bf Qi Chwen Ong$^{1}$ \and Duy Phung$^{1}$ \and Mathieu Ravaut$^{1,2}$ \and} \\ {\bf Shafiq Joty$^{1,3}$ \and Josip Car$^{1,4}$} \\
$^1${Nanyang Technological University, Singapore} \\
$^2${Institute of Infocomm Research (I$^{2}$R), A$^{*}$STAR, Singapore} \\
$^3${Salesforce Research, USA} \\
$^4${Imperial College London, United Kingdom} \\
}

\begin{document}
\maketitle

\begin{abstract}
Low-quality data can cause downstream problems in high-stakes applications. Data-centric approach emphasizes on improving dataset quality to enhance model performance. High-quality datasets are needed for general-purpose Large Language Models (LLMs) training, as well as for domain-specific models, which are usually small in size as it is costly to engage a large number of domain experts for their creation. Thus, it is vital to ensure high-quality domain-specific training data. In this paper, we propose a framework for enhancing the data quality of original datasets\footnote{Code and dataset are available at \\ \url{https://github.com/IvaBojic/framework}}. We applied the proposed framework to four biomedical datasets and showed relative improvement of up to 33\%/40\% for fine-tuning of retrieval/reader models on the \textit{BioASQ} dataset when using back translation to enhance the original dataset quality.
\end{abstract}

\section{Introduction}
\textit{Data-centric} approach emphasizes the collection of high-quality data as a centrally important step in the model development \cite{jarrahi2022principles}. While model-centric approaches were more prominent in the past, recently data-centric approaches are also gaining importance \cite{xu2021dataclue, liu2021data}. This trend was especially emphasized since 2021 when Andrew Ng launched his campaign for a more data-centric approach to AI by starting the data-centric competition\footnote{\url{https://https-deeplearning-ai.github.io/data-centric-comp}}, which encouraged participants to increase accuracy by solely improving the datasets while keeping the model fixed.

Large Language Models (LLMs), such as Generative Pre-trained Transformer 3 (GPT-3) \cite{floridi2020gpt}, generate text that is grammatically correct, fluent, and informative. However, there is little to no control over the data that were used for model training. Consequently, LLMs are prone to hallucinating and providing untruthful outputs \cite{evans2021truthful}. Ironically, this reflects LLMs' ability to be better at learning the training distribution and consequently follow inverse scaling law \cite{lin2021truthfulqa}. And while some of the recent research efforts are focused on providing explanations of where the LLM's outputs came from \cite{menick2022teaching}, such research is in its infancy. 

In this work, we focus on language models with a Transformer encoder architecture such as \textit{BERT} \cite{devlin2018bert}, that extract relevant outputs from a domain-specific evidence-based text corpus. Deep neural networks trained on domain-specific datasets, including those used in Natural Language Processing (NLP), are most heavily dependent on the quality of the training dataset, which is usually small in size \cite{zarcone2021small} as it is costly to engage a large number of domain experts for annotation. It is thus important to create high-quality training data for language models to perform better. In this paper, we propose a data-centric framework for Machine Reading Comprehension (MRC) datasets that increases the original dataset quality by both (i) keeping the size of the original dataset fixed, and (ii) augmenting the original dataset by adding new training samples.

MRC is a Natural Language Understanding (NLU) task. Its goal is to answer questions based on the information provided in a passage \cite{zhang2020machine}. Training datasets for MRC models come in the form of triplets: passage (i.e., positive context), question, and answer. Typically, the MRC pipeline works in two phases, where a passage \textit{retriever} is followed by a passage \textit{reader} \citep{chen2017reading}. For a given question, the retriever first extracts a set of relevant passages from a knowledge base (i.e.,\ text corpus), and then the reader selects an answer (e.g.,\ text span) from one of the retrieved passages \citep{zhu2021retrieving}.

\section{Related Work}
Data-centric approaches can be divided into (i) \textit{data quality enhancement methods} that keep the original size of the dataset fixed (e.g., data filtering or label consistency checking), and (ii) \textit{data augmentation methods} that increase the original dataset size (i.e., adding more training samples). Results from the literature on using data-centric approaches to improve model performance in MRC are inconclusive. 

Several studies have reported that data filtering can lead to significant model improvements \cite{dou2020dynamic, sanyal2021fix, molla2022query}. However, this might not hold if data are filtered in a random way \cite{victoria2021advantages}. Additionally, while increasing labelling consistency and excluding or cleaning noisy data points were shown to improve model performance on the \textit{BioASQ} dataset \cite{yoon2022data}, shortening answers in \textit{AASDQA} led to a decrease of F1-score by 4\% \cite{victoria2021advantages}. 

Adaptation of data augmentation is still comparatively less explored in NLP \cite{feng2021survey}, with a body of work presenting positive results \cite{kaushik2019learning, khashabi2020more, qin2020cosda, pappas2022data} as well as papers showing little or no improvements for the given task \cite{huang2020counterfactually, chopard2021learning, okimura2022impact}.

To the best of our knowledge, this paper is the first that proposes framework for data quality enhancement for improving domain-specific MRC datasets by (i) keeping the original dataset size of data the same and (ii) increasing the original dataset size using augmentation methods. Our framework includes methods for (i) a better selection of negative passages for retriever training, and (ii) reformulating questions using paraphrasing, word substitution, and back translation. 

Paraphrasing, word substitution, and back translation were previously used as data augmentation methods in various NLP tasks \cite{liu2021backtranslation, pappas2022data, ishii2022can}. However, those papers did not look at how each of these methods enhanced the original dataset without increasing its size.  Keeping the size of the dataset fixed is necessary in resources-constrained scenarios, as the resources (e.g., time) needed for fine-tuning are proportional to the size of training sets. Moreover, previous studies did not present a cost-benefit analysis of the resources needed to generate extended training sets and perform fine-tuning processes in comparison with the performance increase.

\section{A Data-centric Framework for MRC}
In our framework, we first generate new training sets using four data quality enhancement methods. We then fine-tune retrieval and reader models on each new training set individually. Secondly, we fine-tune retrieval/reader models continually starting from the best individual checkpoint using enhanced training sets that showed improvements in the first step. Finally, we create new augmented datasets by concatenating all training sets if they show fine-tuning improvements in the first step.

Labels in MRC datasets are triplets which include a passage, a question, and an answer. In MRC datasets, an answer is part of a passage which is also called a \textit{positive context}. To fine-tune a retrieval model as proposed in \cite{karpukhin-etal-2020-dense}, it is necessary to not only provide a positive context of passages that contains the answer to a given question, but also \textit{negative contexts}. Some previous work employed a method of randomly selecting negative contexts from a text corpus \cite{bojic2022sleepqa}. Here we propose a method to improve the random selection of negative contexts. 

One of the problems with manually collecting labels for MRC datasets is that questions are too similar to their answers \cite{rajpurkar2018know}. To solve this, we investigate the use of three different methods applied to the original set of questions: (i) \textit{paraphrasing} - we use two different language models fine-tuned for paraphrasing; (ii) \textit{word substitution} - we use two libraries: one to extract a keyword from a given question and another to obtain a list of synonyms for the chosen keyword; and (iii) \textit{back translation} - we use 25 different machine translation language models to translate a source text into another language, and back into the original language.

\subsection{Negative Contexts}

To enhance the quality of the negative contexts for each passage, we implemented the following procedure. For each positive context, passages were sorted in ascending order of BERTScore \cite{zhang2019bertscore} similarity with the positive context, and the ones with the lowest score were kept to form negative contexts. A global counting dictionary was maintained to prevent the replication of negative contexts across different training examples. This ensured that each negative context did not exceed the \textit{threshold} for number of occurrences in total in the whole dataset.

\subsection{Questions}

In this section, we describe the various techniques used to augment the questions from MRC datasets.

For \textit{question paraphrasing}, we used two models: \textit{T5}\footnote{\url{https://huggingface.co/Vamsi/T5_Paraphrase_Paws}} and \textit{Pegasus}\footnote{\url{https://huggingface.co/tuner007/pegasus_paraphrase}}. To enhance the data quality of an original dataset, for each original question, we used the two aforementioned methods to generate up to five paraphrased questions. Subsequently, we created five different training sets in which we grouped the most, second most, up to the least similar paraphrases for each original question together. The word similarity was calculated using a word vector model from \textit{spaCy}\footnote{\url{https://spacy.io/models/en##en_core_web_lg}}. We also generated a sixth set comprising a randomly-selected question from the list of five unique paraphrases generated.

In \textit{word substitution} process, we extracted a keyword from each question with the help of the \textit{spaCy} library and obtained a list of synonyms for each keyword using Natural Language Tool Kit (NLTK)'s English dictionary, \textit{WordNet}\footnote{\url{https://www.nltk.org/howto/wordnet.html}}. The top five synonyms were extracted from this list in descending order of word similarity calculated using the aforementioned word vector model from \textit{spaCy}. We then generated five versions of the training data for each dataset such that in set 1, the keyword for each question was replaced by its most similar synonym; in set 2, the keyword for each question was replaced by its second most similar synonym and so forth, with set 5 containing the questions with the least similar synonyms as substitutes. For keywords with \textit{n} < 5 synonyms, we kept the question unchanged in the first (5 - \textit{n}) versions and used the synonyms as substitutes in the remaining \textit{n} versions. We also created a sixth set in which we randomly selected one of the top five (or \textit{n}) synonyms to substitute the keyword for each question.

We used \textit{Hugging Face Helsinki} model\footnote{\url{https://huggingface.co/Helsinki-NLP}} for \textit{back translation}. In total, we generated 25 different training sets based on the number of downloads for translation from English to the respective languages, followed by the availability of translation models from the respective languages to English. We selected checkpoints based on the number of downloads, taking the top 25 most downloaded. 

To understand how different the resulting questions obtained from each of the enhancement methods are, we performed pairwise comparisons between questions from each method using ROUGE-1. Results are shown in \Cref{subsec:b7}. \emph{Back-translation} overall yields the questions most different to the baseline and the other enhancement methods. 

\subsection{Answers}

Since MRC relies on extracting the exact answer (i.e., text span) from a passage, we could not apply any of the automatic data quality enhancement methods that we applied to questions (as explained in the previous section). However, we created new training datasets in which we manually shortened the original answers wherever appropriate. We explained further in \Cref{subsec:A3}.

\section{Datasets}
To test our framework, we made adjustments (see Appendix \ref{sec:appendix1}) to four biomedical datasets: \textit{BioASQ} \cite{lamurias2020generating}, \textit{COVID-QA} \cite{moller2020covid}, \textit{cpgQA} \cite{mahbub2023cpgqa} and \textit{SleepQA} \cite{bojic2022sleepqa}. We refer the reader to \Cref{tab1} for statistics on the final version of datasets that we used in all experiments: original/final size of text corpus, original/final number of labels and finally, train/dev/test split.

Original \textit{BioASQ} dataset contained over 3k manually-annotated biomedical labels. Questions in these datasets came in four different flavours: factoid, list, yes/no, and summary. We extracted only factoid questions for which the exact answer can be found in the positive context. Original \textit{COVID-QA} dataset was annotated by biomedical experts and contained 2k labels on COVID-19 pandemic-related topics. Original \textit{cpgQA} dataset contained 1k manually annotated labels in the domain of clinical practice guidelines. Original \textit{SleepQA} was a manually annotated dataset in the sleep domain with 5k labels. 

\begin{table}[h!]
\centering
\caption{Dataset statistics, for original and final versions.}
\resizebox{\columnwidth}{!}{
\begin{tabular}{l|rrrrr}
\toprule
\textbf{Dataset} 
& \textbf{\begin{tabular}[r]{@{}r@{}}Original \\  corpus \end{tabular}} 
& \textbf{\begin{tabular}[r]{@{}r@{}}Final \\  corpus\end{tabular}} 
& \textbf{\begin{tabular}[r]{@{}r@{}}Original \\  labels \end{tabular}} 
& \textbf{\begin{tabular}[r]{@{}r@{}}Final \\  labels \end{tabular}} 
& \textbf{\begin{tabular}[r]{@{}r@{}}Final \\  train/dev/test \end{tabular}}  \\
\hline
\textbf{\textit{BioASQ}} & 4265 & 957 & 5821 & 957 & 765/96/96\\
\textbf{\textit{COVID-QA}} & 2079 & 1121 & 1327 & 1102 & 896/112/113 \\
\textbf{\textit{cpgQA}} & 190 & 235 & 1097 & 1097 & 877/110/110 \\
\textbf{\textit{SleepQA}} & 7000 & 7000 & 5000 & 5000 & 4000/500/500 \\
\bottomrule
\end{tabular}
}
\label{tab1}
\end{table}

\section{Evaluation}
We evaluated our framework by performing fine-tuning of retrieval and reader models using \textit{BioLinkBERT} \cite{yasunaga2022linkbert} and \textit{BioBERT BioASQ}\footnote{\url{https://huggingface.co/gdario/biobert_bioasq}} respectively. We used \textit{BioLinkBERT} for retrieval model fine-tuning as it was recently shown to achieve state-of-the-art performance on low-resource bio-MRC tasks \cite{mahbub2023cpgqa}. \textit{BioBERT BioASQ} was used for fine-tuning of reader model as proposed in \cite{bojic2022sleepqa}. Intrinsic evaluation of fine-tuned models was done using automatic metrics on test sets: recall@1 for retrieval and Exact Match (EM) for reader models.

\subsection{Fine-tuning on Enhanced Training Sets}

\Cref{tab2} and \Cref{tab3} show recall@1/EM scores respectively for fine-tuned retrieval/reader models after enhancing the method of selecting negative contexts (i.e., using \textit{BertScore}) for the retrieval training datasets, as well as reformulation of questions using paraphrasing, word substitution, back translation and answer shortening for the training datasets of \emph{both} models. More specifically: 

\begin{itemize}[leftmargin=*]

    \item The first row (\emph{baseline}) in each table shows the results of \textit{BioLinkBERT}/\textit{BioBERT BioASQ} models fine-tuned on the original datasets.

    \item Each subsequential row shows the best results for each dataset using the four aforementioned methods for negative contexts (only for the retrieval models) and questions (for both models) enhancement. 

    \item The following row (\emph{answer shortening}) shows recall@1/EM scores for fine-tuning of models on the training datasets in which the original answers were manually shortened if needed. 

    \item The following row (\emph{continual}) shows the results of \emph{continual fine-tuning}: starting from the best individual checkpoint, we fine-tune on the second-best training set, and so on. For example, for reader fine-tuning on the \textit{BioASQ} dataset, we first took the checkpoint of fine-tuning on the training set created using paraphrasing and then continued fine-tuning on training sets created using back translation. Finally, we took the newest checkpoint and continued fine-tuning on the training set created using word substitution.

    \item The last row (\emph{augmentation}) shows recall@1/EM scores for fine-tuning of models on the training datasets created by concatenating all data enhanced training sets if they showed fine-tuning improvements when using individually (i.e., rows 2-6 for retrieval and rows 2-5 for reader models).
    
\end{itemize}

For retrieval fine-tuning (\Cref{tab2}), the most significant improvement of +8.3 (+33\%) from baseline was achieved for \textit{BioASQ} dataset when using back translation on the Catalan language. The enhanced methods of selecting negative contexts and word substitution improved all four datasets, while paraphrasing and back translation caused a decrease in recall@1 scores for \textit{SleepQA} dataset. Continual retrieval fine-tuning showed improvements when compared to baselines for all datasets, however, only for the \textit{COVID-QA} and \textit{cpgQA} datasets it was better than the best results of individual fine-tuning. 

\begin{table}[h!]
\centering
\caption{Results of fine-tuned \textbf{retrieval} models (recall@1).}
\resizebox{\columnwidth}{!}{
\begin{tabular}{l|cccc}
\toprule
\textbf{Methods} & \textit{\textbf{BioASQ}} & \textit{\textbf{COVID-QA}} & \textit{\textbf{cpgQA}} & \textit{\textbf{SleepQA}} \\
\hline
\textbf{baseline} & \textbf{25.0} & \textbf{42.5} & \textbf{66.4} & \textbf{46.8} \\
\hline
\textbf{negatives} & \cellcolor{green!50}32.3 & \cellcolor{green!10}48.7 & \cellcolor{green!10}67.3 & \textbf{\cellcolor{green!100}48.4} \\
\textbf{paraphrasing} & \cellcolor{green!10}31.2 & \cellcolor{green!80}54.0 & 66.4 & \cellcolor{red!30}46.6 \\
\textbf{word substitution} & \cellcolor{green!30}30.2 & \cellcolor{green!50}50.4 & \cellcolor{green!50}69.1 & \textbf{\cellcolor{green!100}48.4} \\
\textbf{back translation} & \textbf{\cellcolor{green!100}33.3} & \cellcolor{green!30}49.6 & 66.4 & \cellcolor{red!50}45.8 \\
\textbf{answer shortening} & \cellcolor{green!10}29.2 & \cellcolor{green!10}45.1 & 66.4 & \cellcolor{red!80}44.8 \\
\hline
\textbf{continual} & \cellcolor{green!5}29.2 & \textbf{\cellcolor{green!100}62.8} & \textbf{\cellcolor{green!100}70.9} & \cellcolor{green!30}47.2 \\
\hline
\textbf{augmentation} & \cellcolor{green!10}31.2 & \cellcolor{green!80}60.2 & \cellcolor{red!30}65.5 & \cellcolor{red!50}45.0 \\
\bottomrule
\end{tabular}
}
\label{tab2}
\end{table}

For fine-tuned reader models (\Cref{tab3}), the most significant improvement of 2.1 (+40\%) from baseline was achieved for \textit{BioASQ} dataset when using back translation on the Dutch language, as well as paraphrasing. Continual reader fine-tuning increased the EM score only for \textit{cpgQA} dataset compared with the corresponding baselines. Lastly, augmentation was better than the best results of individual fine-tuning only for the \textit{SleepQA} dataset with the total increase of 2.6 (+4\%). 

\begin{table}[h!]
\centering
\caption{Results of fine-tuned \textbf{reader} models (EM). *Since no single enhancement method could improve the baseline on cpqQA, we discarded \emph{continual} and \emph{augmentation} on this dataset.}
\resizebox{\columnwidth}{!}{
\begin{tabular}{l|cccc}
\toprule
\textbf{Methods} & \textit{\textbf{BioASQ}} & \textit{\textbf{COVID-QA}} & \textit{\textbf{cpgQA}} & \textit{\textbf{SleepQA}} \\
\hline
\textbf{baseline} & \textbf{5.2} & \textbf{22.1} & \textbf{50.9} & \textbf{58.6} \\
\hline
\textbf{paraphrasing} & \textbf{\cellcolor{green!100}7.3} & \textbf{\cellcolor{green!100}23.9} & \textbf{50.9} & \cellcolor{green!10}{59.0} \\
\textbf{word substitution} & \cellcolor{green!30}{6.3} & 22.1 & \textbf{50.9} & \cellcolor{green!30}{59.4} \\
\textbf{back translation} & \textbf{\cellcolor{green!100}7.3} & \cellcolor{green!30}23.0 & \cellcolor{red!50}46.4 & \cellcolor{green!30}{59.4} \\
\textbf{answer shortening} & 5.2 & \cellcolor{green!30}23.0 &  \cellcolor{red!50}{49.1} & \cellcolor{green!80}{60.8} \\
\hline
\textbf{continual} & 5.2 & \textbf{\cellcolor{green!100}23.9} & N/A* & \cellcolor{red!50} 58.0 \\
\hline
\textbf{augmentation} & 5.2 & \textbf{\cellcolor{green!100}23.9} & N/A* & \textbf{\cellcolor{green!100}{61.2}} \\
\bottomrule
\end{tabular}
}
\label{tab3}
\end{table}

Greater relative improvements with \emph{back-translation} compared to other methods could be supported by this method creating more diverse questions (\Cref{subsec:b7}). However, \emph{back-translation} gains are inconsistent from a dataset to the other. Moreover, we noticed that translation and paraphrasing with \textit{Pegasus} gave questions noticeably more difference than the other data enhancing techniques.

\subsection{Cost-benefit Analysis}

In total, the data-centric methods that we described previously enabled us to generate 28 and 24 enhanced training sets for retrieval fine-tuning and reader fine-tuning respectively. Subsequently, we fine-tuned all retrieval/reader models on a single NVIDIA-A40 GPU with 46GB of GPU RAM. \Cref{tab4} and \Cref{tab5} shows time spent on fine-tuning. For example, we used one GPU for five hours to fine-tune retriever model on \textit{BioASQ} dataset to achieve 33\% improvement in recall@1 score. Meanwhile, we used one GPU for 22 hours to fine-tune retriever model on \textit{SleepQA} dataset only to achieve a decrease in recall@1 score of 2\%. 

The last two rows in tables show the time needed for continual/augmentation fine-tuning only. However, in order to determine the order in which to fine-tune for continual learning, or which datasets to use for concatenation, all individual checkpoints need to be created. Hence, to obtain the total time for continual learning/augmentation, one needs to add up times from all previous rows as well.

\begin{table}[h!]
\centering
\caption{Total time spent (in hours) vs.\ maximum relative recall@1 improvements of retrieval fine-tuning.}
\resizebox{\columnwidth}{!}{
\begin{tabular}{l|rrrr}
\toprule
\textbf{Methods} & \textit{\textbf{BioASQ}} & \textit{\textbf{COVID-QA}} & \textit{\textbf{cpgQA}} & \textit{\textbf{SleepQA}} \\
\hline
\textbf{baseline} & \textbf{0.2} & \textbf{0.2}  & \textbf{0.2}  & \textbf{0.9}   \\
\hline
\textbf{negatives} &\cellcolor{green!50} 0.9 (29\%) & \cellcolor{green!10}1.1 (15\%) & \cellcolor{green!10} 1.0 (1\%) & \cellcolor{green!100}9.9 (3\%)\\
\textbf{paraphrasing} & \cellcolor{green!10}4.3 (25\%) & \cellcolor{green!80}3.7 (27\%) & 3.6 (0\%) & \cellcolor{red!10}25.4 (1\%) \\
\textbf{substitution} & \cellcolor{green!30} 2.5 (21\%) & \cellcolor{green!50}1.4 (19\%) & \cellcolor{green!50}1.8 (4\%) & \cellcolor{green!100} 6.1 (3\%)\\
\textbf{translation} & \cellcolor{green!100} 4.9 (33\%) & \cellcolor{green!30} 6.3 (17\%) & 4.9 (0\%) & \cellcolor{red!50} 22.0 (2\%) \\
\textbf{answer shortening} & \cellcolor{green!10} 0.4 (17\%) & \cellcolor{green!10} 0.4 (6\%) &  0.4 (0\%) & \cellcolor{red!80} 1.6 (4\%) \\
\hline
\textbf{continual} & \cellcolor{green!5} 1.6 (17\%) & \cellcolor{green!100} 1.7 (48\%) & \cellcolor{green!100} 0.7 (7\%) & \cellcolor{green!30} 1.1 (1\%) \\
\hline
\textbf{augmentation} & \cellcolor{green!10}0.9 (25\%) & \cellcolor{green!80}1.0 (42\%) & \cellcolor{red!30} 0.6 (1\%) & \cellcolor{red!50} 2.6 (4\%) \\
\bottomrule
\end{tabular}
}
\label{tab4}
\end{table}

\begin{table}[h!]
\centering
\caption{Total time spent (in hours) vs.\ maximum relative EM improvements of reader fine-tuning.}
\resizebox{\columnwidth}{!}{
\begin{tabular}{l|rrrr}
\toprule
\textbf{Methods} & \textit{\textbf{BioASQ}} & \textit{\textbf{COVID-QA}} & \textit{\textbf{cpgQA}} & \textit{\textbf{SleepQA}} \\
\hline
\textbf{baseline} & \textbf{0.1} & \textbf{0.1} & \textbf{0.1} & \textbf{0.3}  \\
\hline
\textbf{paraphrasing}&\cellcolor{green!100} 1.0 (40\%) & \cellcolor{green!100} 0.9 (8\%) & 0.9 (0\%) & \cellcolor{green!10}5.0 (1\%) \\
\textbf{substitution} & \cellcolor{green!30}0.3 (21\%) & 0.5 (0\%) & 0.3 (0\%) & \cellcolor{green!30}1.1 (1\%) \\
\textbf{translation}& \cellcolor{green!100}1.0 (40\%) & \cellcolor{green!30}1.2 (4\%) & \cellcolor{red!50}1.7 (9\%) & \cellcolor{green!30}4.0 (1\%) \\
\textbf{answer shortening} & 0.1 (0\%) & \cellcolor{green!30} 0.1 (4\%) &  \cellcolor{red!50}{0.1 (4\%)} & \cellcolor{green!80}{0.2 (4\%)} \\
\hline
\textbf{continual} & 0.2 (0\%) & \cellcolor{green!100}0.3 (8\%) & N/A & \cellcolor{red!50} 1.4 (1\%) \\
\hline
\textbf{augmentation} & 0.1 (0\%) & \cellcolor{green!100} 0.1 (8\%) & N/A & \cellcolor{green!100} 0.5 (4\%)  \\
\bottomrule
\end{tabular}
}
\label{tab5}
\end{table}

\section{Discussion and Conclusions}
It is estimated that over 92\% of data scientists who work in the Artificial Intelligence field encountered the “data cascades” phenomenon, which denotes downstream problems resulting from low-quality data \cite{sambasivan2021everyone}. One way to improve the original dataset quality is data-centric approach. In this paper, we showed that by enhancing the quality of original datasets, one can achieve model fine-tuning performance improvements for small datasets (e.g., biomedical datasets). However, the results suggest that the effects of data quality enhancement methods on performance improvements are small, and the performance of the test data deteriorates in many cases.
 
Despite the inconsistency of data-centric methods used in this paper in yielding improvement, two positive conclusions can be drawn. First, when taking into consideration the time needed to create data enhanced training sets as well as performance improvements in fine-tuning, word substitution method is the best, supporting previous findings \cite{feng2019keep, pappas2022data}. Unlike other methods, word substitution is not model-based and thus is run for a few minutes, rather than a few hours like back translation and paraphrasing. Second, the best relative improvements were achieved for the \textit{BioASQ} dataset with the smallest number of labels, a similar finding presented in \cite{okimura2022impact}.

In addition to the data-centric methods discussed in this work, there are other techniques such as pseudo-labelling \cite{abney2007semisupervised, ruder2018strong, cui2019self, zhu2022introduction}, data selection \cite{axelrod2011domain, plank2011effective, ruder2017learning}, and pre-training methods \cite{han2019unsupervised, guo2020multi}. In future work, we will investigate whether those techniques would produce more reliable and consistent results across different datasets. Moreover, we will also consider approaches that incorporate aspects of multiple techniques, resulting in hybrid data-centric techniques as proposed in \cite{ramponi2020neural}.

\section*{Acknowledgements}

The authors would like to acknowledge the funding support from Nanyang Technological University, Singapore. Josip Car’s post at Imperial College London is supported by the NIHR NW London Applied Research Collaboration.

\clearpage

\balance

\bibliography{references}
\bibliographystyle{acl_natbib}

\appendix

\clearpage

\section{Datasets}
\label{sec:appendix1}

\subsection{Dataset Construction}

In this subsection, we describe how we built the final version of datasets from \Cref{tab1}. Where necessary, we divided passages from the original text corpus into one or more parts, so their length was less than 300 words. This step was done so that all passages were of a similar length across different datasets and that the same model hyperparameters can be used for fine-tuning retrieval and reader models. We then removed those labels for which the answer could not be found in the corresponding positive context. Finally, we divided each original dataset into three parts (in the ratio of 80:10:10) to create training, development, and test sets. \Cref{tab1} shows the original number of passages in each text corpora, the original number of labels, and the final numbers after the aforementioned adjustments were done.

\subsection{Data Cleaning}

\textit{BioASQ}: The original dataset did not include positive passages, but instead contained links to the journal articles where the answers can be found. To obtain positive passages, we first retrieved them from the individual links provided in the dataset, and then divided them into passages of no longer than 300 words. Only triplets that contain the exact answers in the retrieved passages were included in the final dataset. We encountered a challenge that, of the 5,821 triplets of the factoid type identified, only 16\% had the exact answers that could be found in the provided passages.

\textit{COVID-QA}: We first divided the original corpus into passages containing no more than 300 words. We also removed redundant keywords, such as 'introduction:', 'introductions:', 'objective:', 'objectives:', 'conclusion:', 'conclusions:', 'method:', 'methods:', 'background:', 'backgrounds:', 'result:', 'results:', 'result(s):', and 'aim:'. Additionally, we eliminated leading and trailing spaces and changed all letters to lowercase. To ensure dataset accuracy, further manual cleaning was carried out. This includes filling in incomplete words, removing medical abbreviations, and correcting missing brackets such as "()" and "[]".

\textit{cpgQA}: To prepare the text corpus, we partitioned passages into segments of no more than 300 words, resulting in a corpus of 235 passages. Unfortunately, this division caused some answers to be separated from their corresponding positive contexts due to issues such as inaccurate sentence tokenization and answer fragmentation between two adjacent passages. These discrepancies were addressed through manual intervention. It should be noted that no labels were excluded from the original dataset as a result of this cleaning procedure.

\textit{SleepQA} The original dataset already contained passages shorter than 300 words, and all answers were found in their provided passages. We eliminated leading and trailing spaces and changed all letters to lowercase.

\subsection{Shortening Answers}
\label{subsec:A3}

\begin{figure}[h!]
\centering
\includegraphics[width=1\linewidth]{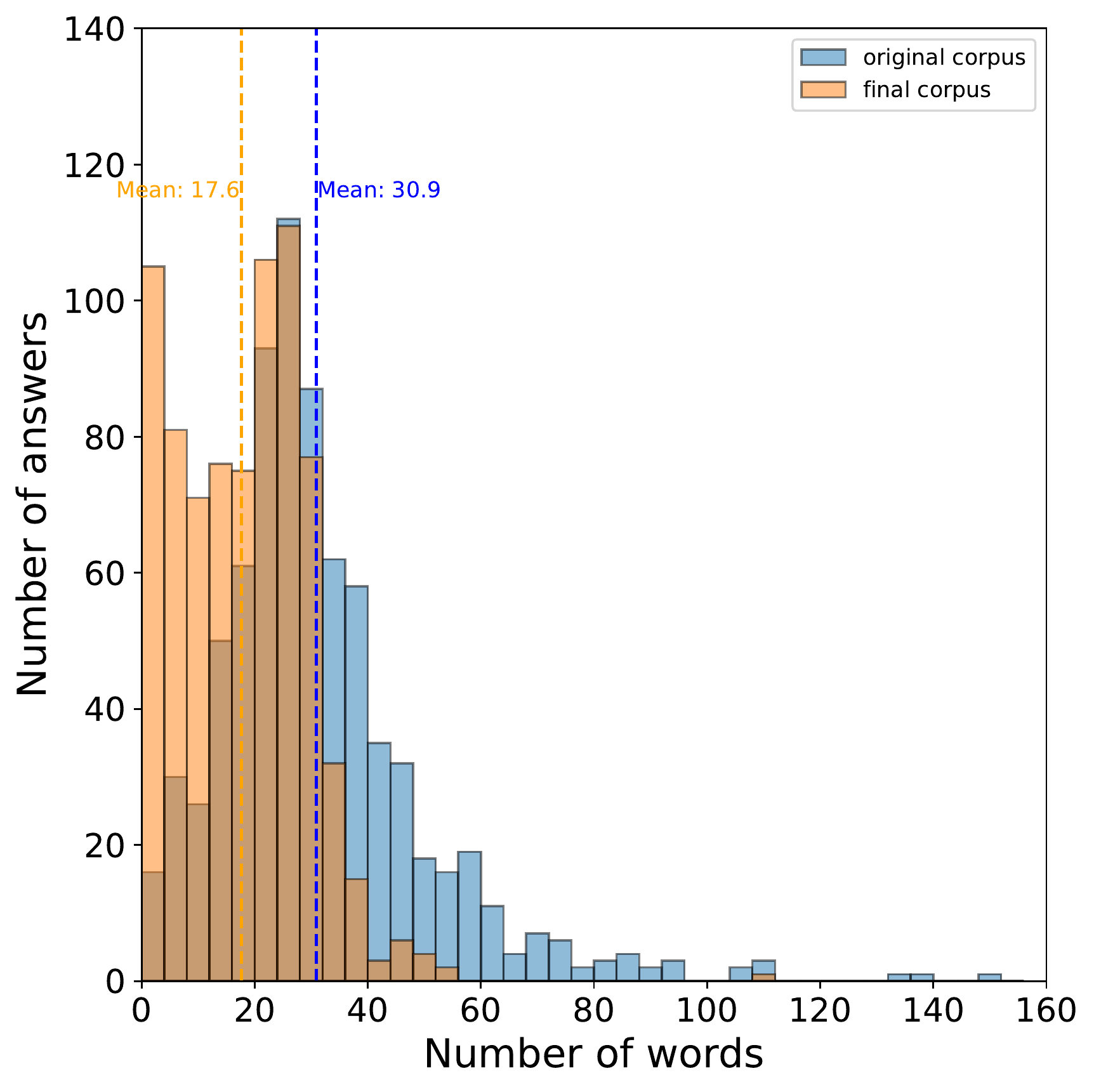}
\caption{Answer length (in number of words) before and after shortening answers for BioASQ training set.}
\label{fig:trainbio}
\end{figure}

\begin{figure}[t!]
\centering
\includegraphics[width=0.92\linewidth]{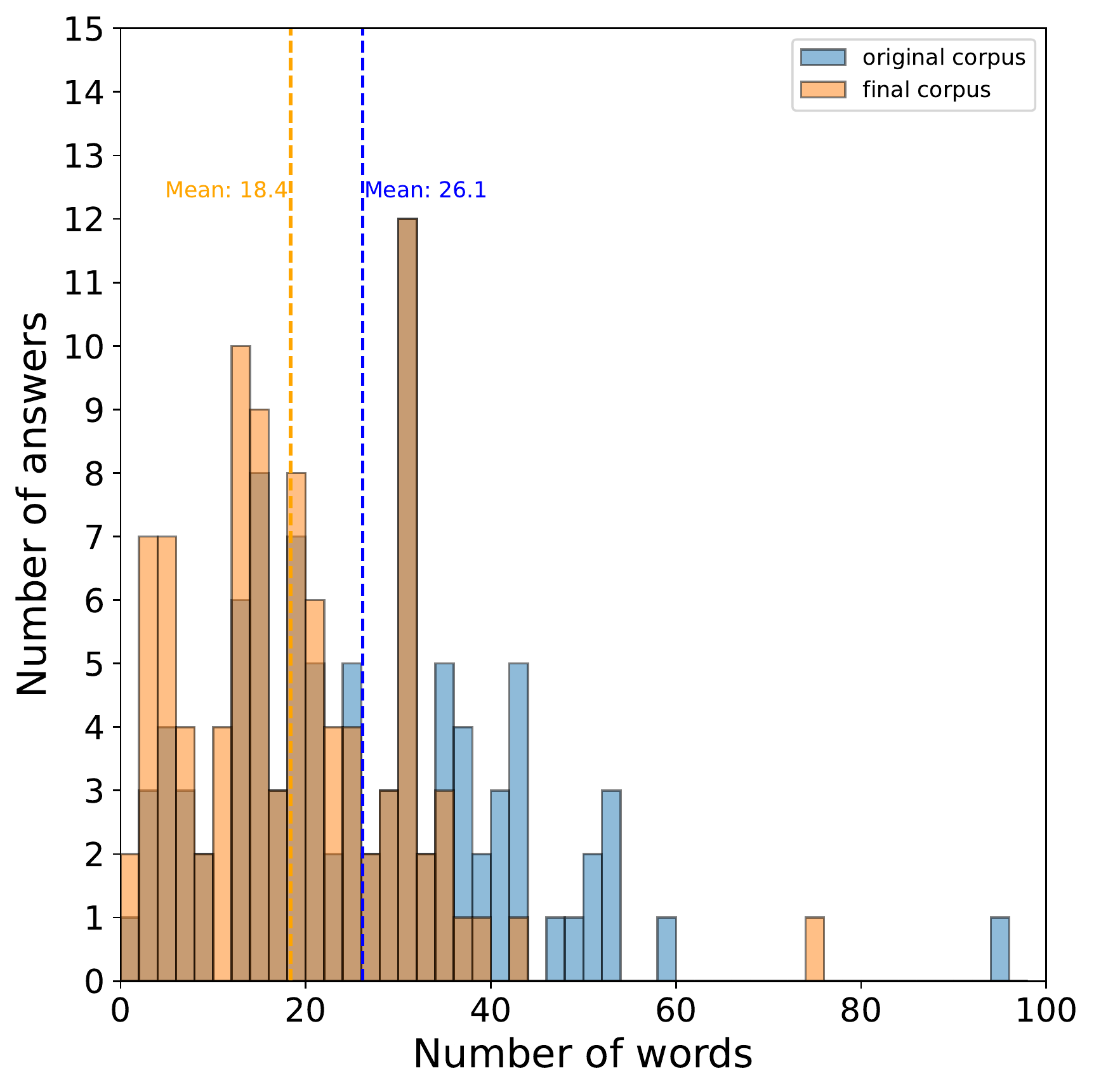}
\caption{Answer length (number of words) before and after shortening answers for the BioASQ test set.}
\label{fig:testbio}
\end{figure}

\textit{BioASQ}: The original answers varied from two to more than 120 words in length. Our focus was on shortening the answers which were excessively long, and thus all answers longer than 30 words were manually reviewed. The primary adjustments made to the answers involved isolating the main response to the corresponding question, thereby truncating lengthy sentences into shorter phrases. This approach effectively reduced answer length for both the test and training sets by a significant degree. The mean answer length for the training set decreased from \textbf{30.9} to \textbf{17.6} words (\Cref{fig:trainbio}), while the mean answer length for the test set decreased from \textbf{26.1} to \textbf{18.4} words (\Cref{fig:testbio}).

\newpage

\textit{COVID-QA}: In the original dataset, the length of the answers was not more than 120 words. However, some answers contained incomplete words at the beginning and/or end of sentences. To improve the dataset's accuracy, these words were either manually removed or completed. Moreover, scientific abbreviations were eliminated manually to improve the accuracy of exact matches. Unfortunately, this had no significant effect on the mean length of answers for both the training and test sets. This result can be attributed to the training set's prevalence of sentences with only one or two abbreviations. In other cases, completing the incomplete words also had no effect on the mean word count.

\renewcommand{\thefigure}{3}
\begin{figure}[h!]
\centering
\includegraphics[width=0.92\linewidth]{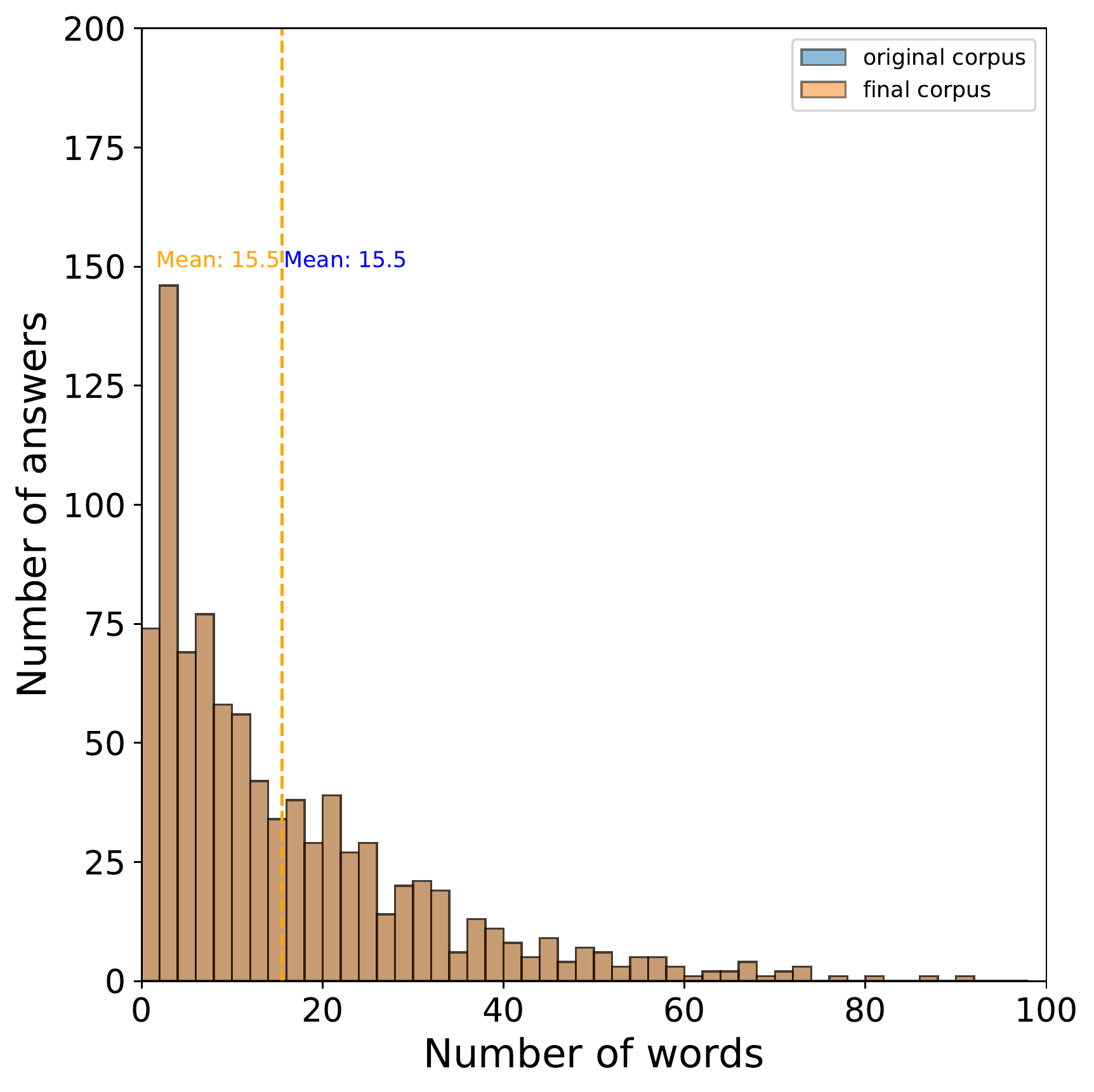}
\caption{Answer length (in number of words) before and after shortening answers for COVID-QA training set.}
\label{fig:traincovid}
\end{figure}

\textit{cpgQA}: In both the training and test sets, answers were shortened manually by removing extraneous phrases and articles (such as "a/an/the") from the beginning of the responses. After shortening, the mean answer length in the training set reduced from \textbf{12.7} words to \textbf{12.4} words, whereas for the test set, the mean answer length reduced from \textbf{12.1} words to \textbf{11.6} words. The minimal difference in the mean number of words is due to the fact that most answers in the original dataset were clear and concise.

\renewcommand{\thefigure}{4}
\begin{figure}[h!]
\centering
\includegraphics[width=0.995\linewidth]{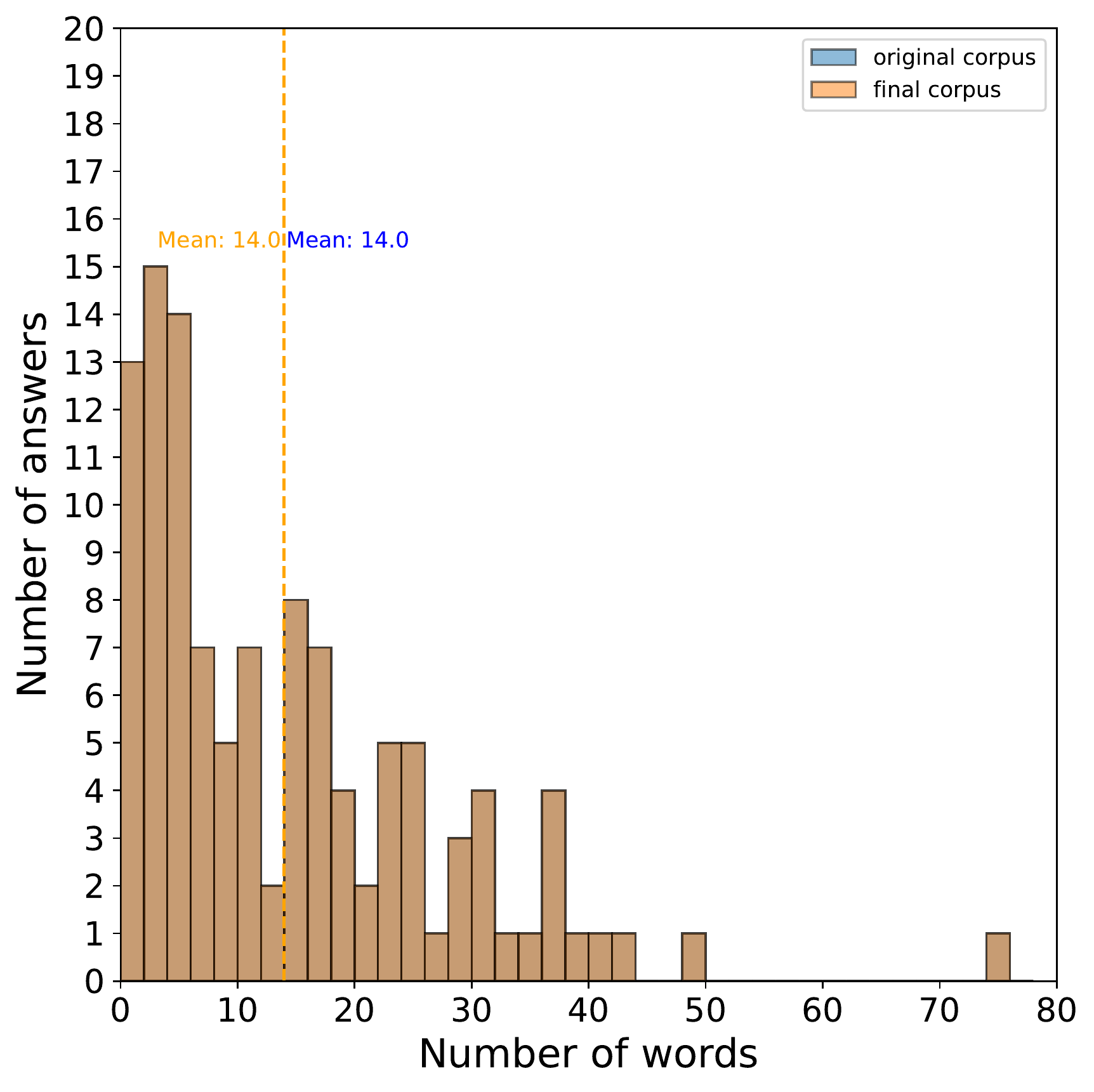}
\caption{Answer length (in number of words) before and after shortening answers for COVID-QA test set.}
\label{fig:testcovid}
\end{figure}

\renewcommand{\thefigure}{5}
\begin{figure}[h!]
\centering
\includegraphics[width=0.995\linewidth]{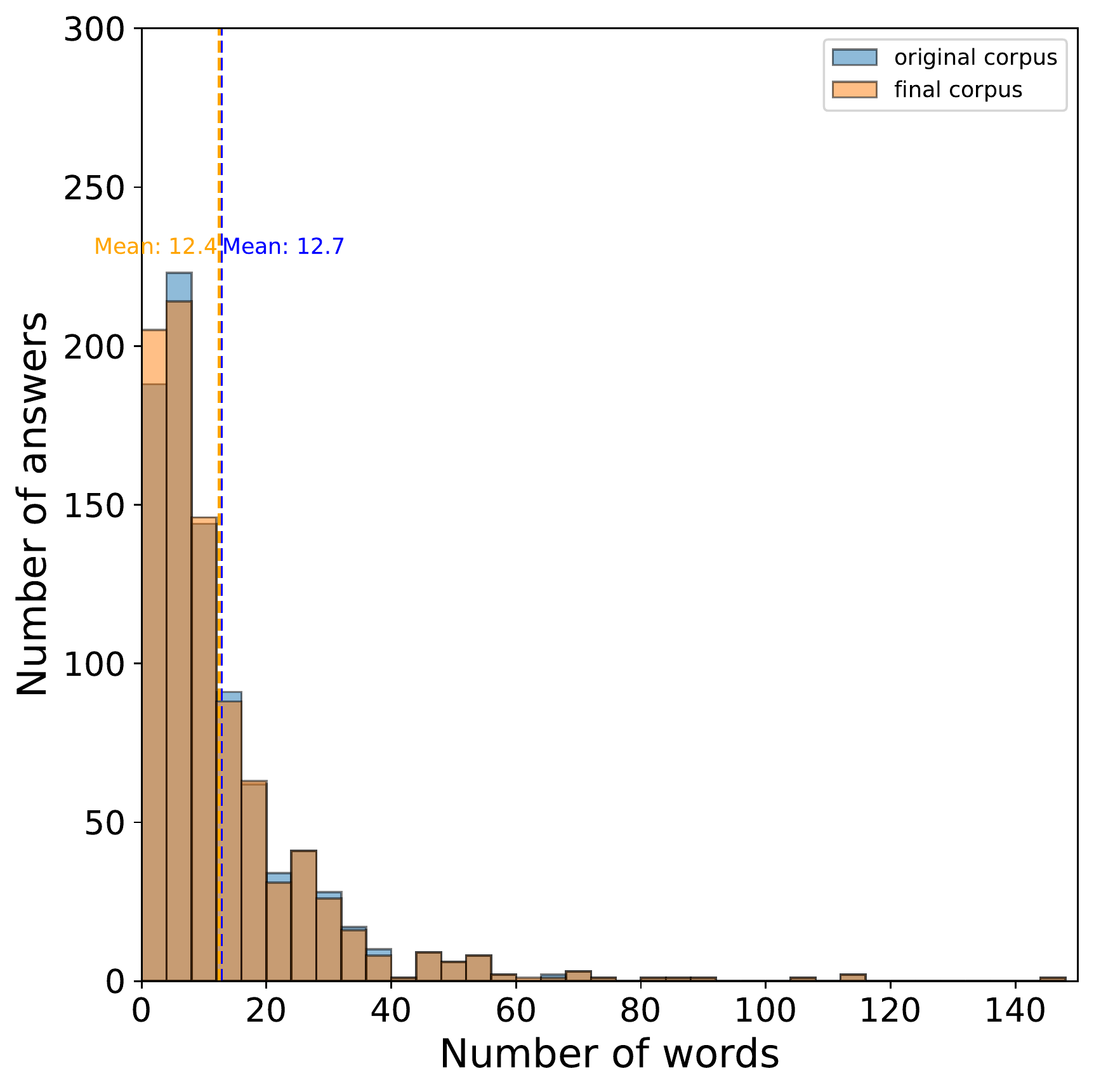}
\caption{Answer length (in number of words) before and after shortening answers for cpgQA training set.}
\label{fig:traincpg}
\end{figure}

\textit{sleepQA}: The initial average answer lengths for the sleepQA dataset are \textbf{10.15} and \textbf{9.13} for the train and test set respectively, making it the dataset with the shortest average answer length among all datasets studied. We focused on cutting down answers more than 15 words long, which range up to 40 words long. The was done by extracting the main phrases of the answers that directly respond to the associated questions. The resulting cleaned answers are in the form of shorter, more concise phrases instead of wordy full sentences. The final average answer lengths after the cleaning process are \textbf{9.11} and \textbf{8.01} for the train and test set respectively.

\renewcommand{\thefigure}{7}
\begin{figure}[h!]
\centering
\includegraphics[width=0.92\linewidth]{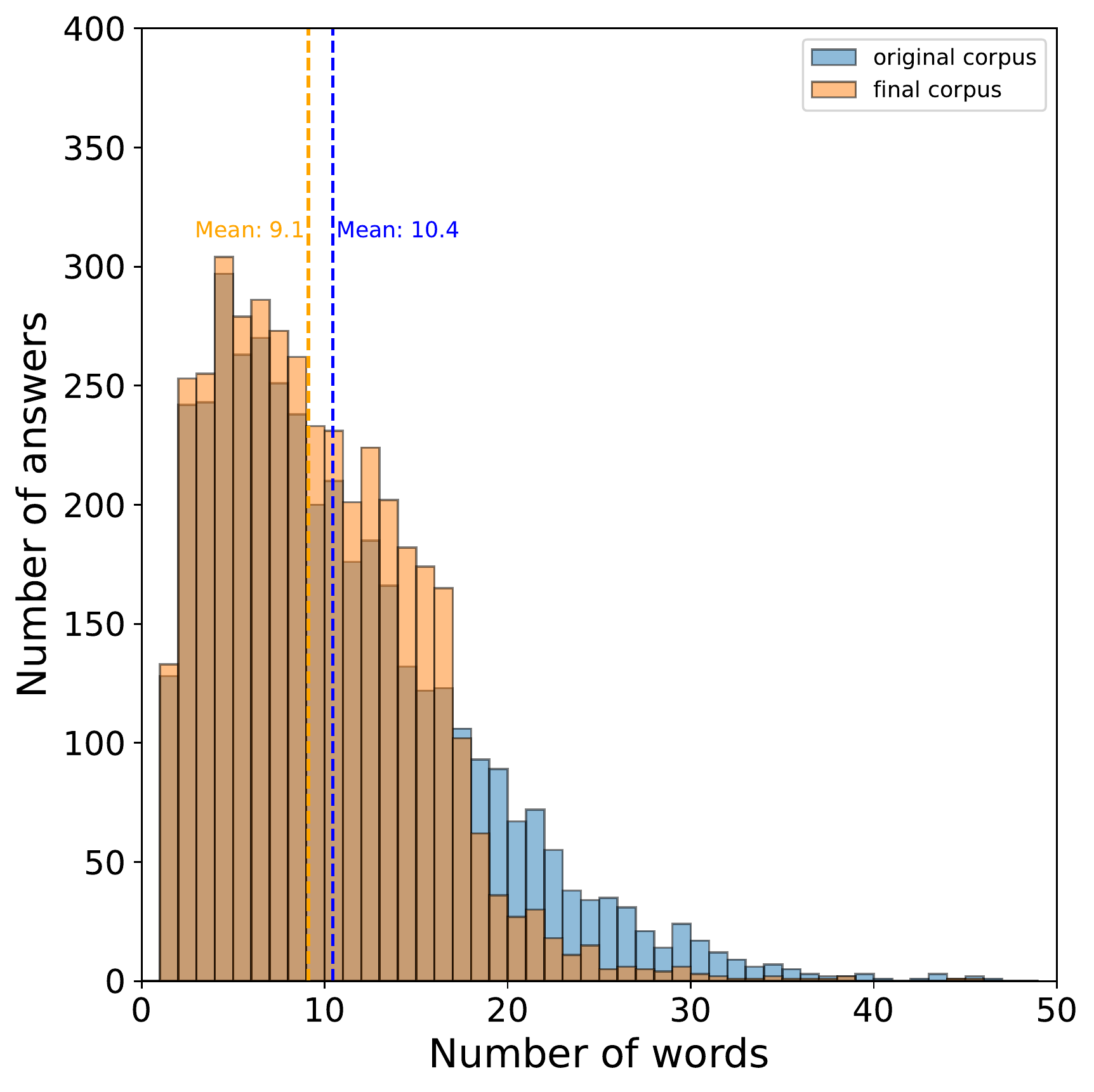}
\caption{Answer length (in number of words) before and after shortening answers for SleepQA training set.}
\label{fig:trainsleep}
\end{figure}

\renewcommand{\thefigure}{8}
\begin{figure}[b!]
\centering
\includegraphics[width=0.85\linewidth]{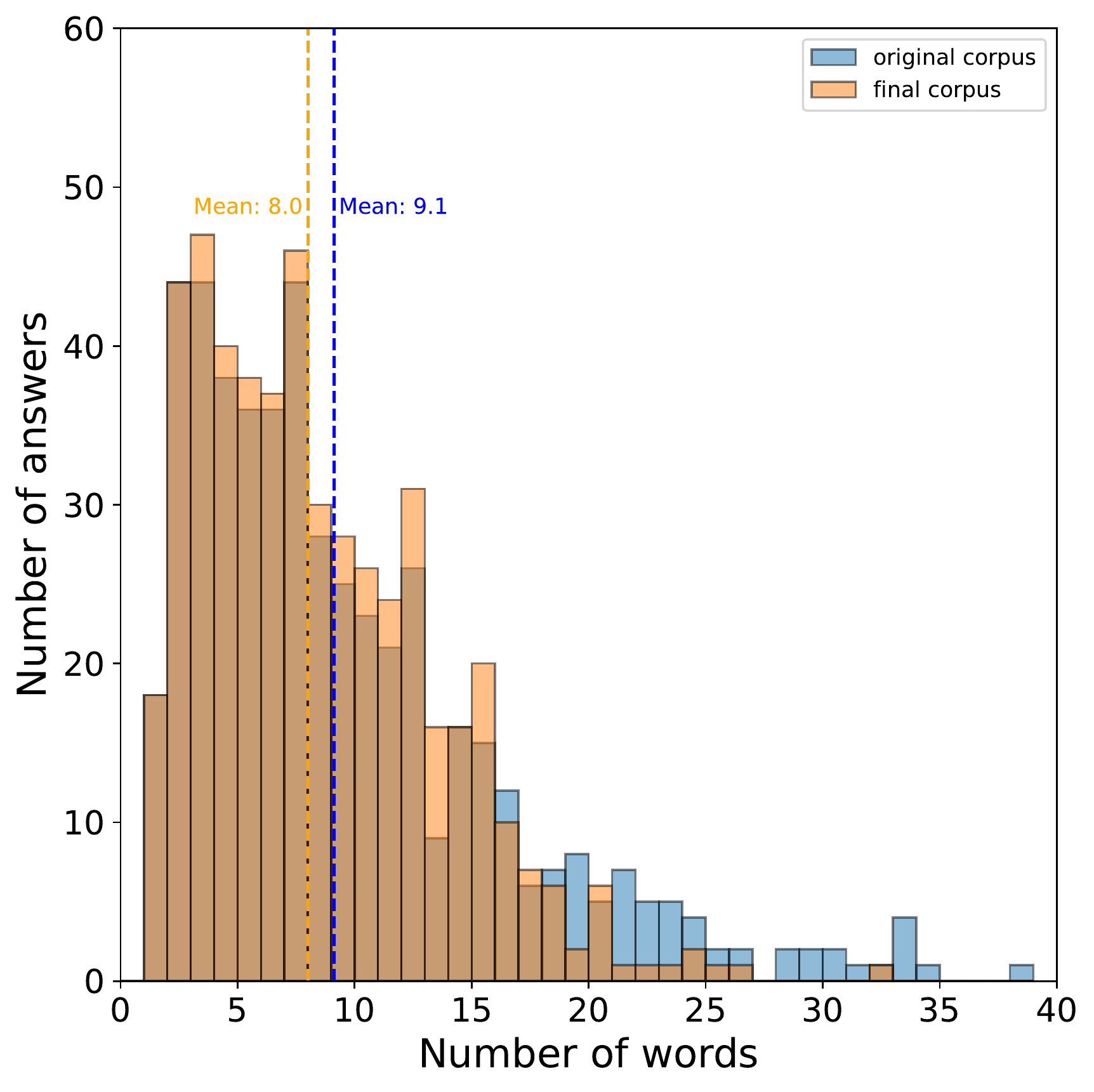}
\caption{Answer length (in number of words) before and after shortening answers for SleepQA test set.}
\label{fig:testsleep}
\end{figure}

\renewcommand{\thefigure}{6}
\begin{figure}[t!]
\centering
\includegraphics[width=0.92\linewidth]{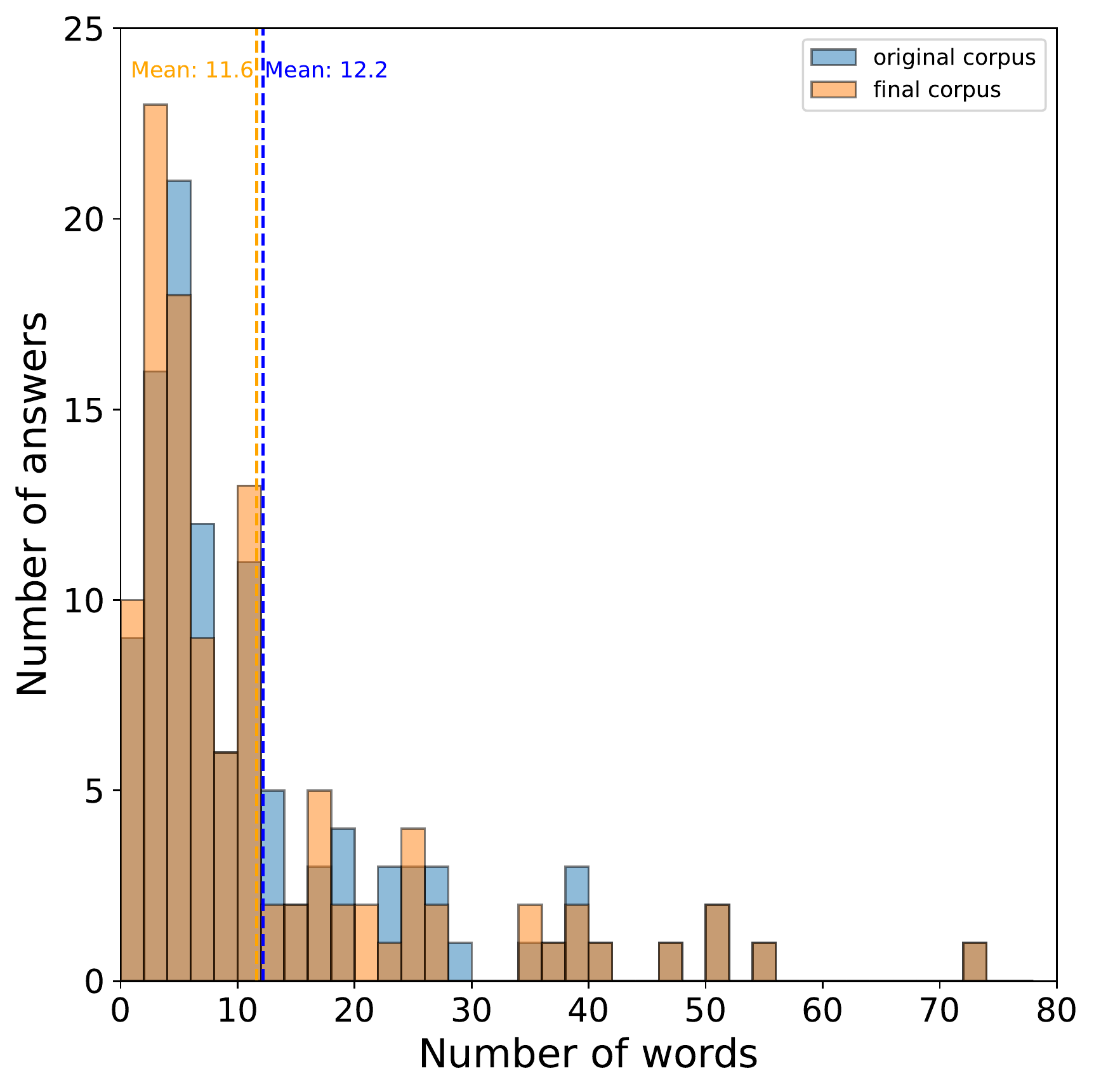}
\caption{Answer length (in number of words) before and after shortening answers for cpgQA test set.}
\label{fig:testcpg}
\end{figure}

\section{Evaluation}
\label{sec:appendix2}

\subsection{Model Hyperparameters}

Hyperparameters of retrieval models fine-tuning are shown in \Cref{tabB01}, and of reader models in \Cref{tabB02}. When fine-tuning retrieval models on training sets in which method of selecting the negative contexts for each passage was enhanced, we changed \textit{other negatives} hyperparameters to reflect the number of negative contexts in the corresponding training set (e.g., 1 to 5). Additionally, when fine-tuning reader models on different datasets, we set \textit{eval step} to 50 for \textit{BioASQ}, \textit{COVID-QA} and \textit{cpgQA} datasets and 500 for the \textit{SleepQA} dataset. The reason behind this is because the \textit{SleepQA} dataset has 4,000 labels in the train set, while the other datasets have less than 1,000 labels. For continual retrieval fine-tuning, we set the \textit{num train epochs} to 60, and for reader to 30. Other parameters were left the same.

\begin{table}[ht!]
 \centering
 \caption{Hyperparameters of retrieval model fine-tuning.}
 \resizebox{0.90\columnwidth}{!}{
 \begin{tabular}{lr}
 \toprule
 \textbf{Hyperparameter} & \textbf{Value} \\
 \midrule
 batch size & 32\\
 dev batch size & 32\\
 adam eps & $1e-8$\\
 adam betas & (0.9, 0.999)\\
 max grad norm & 1.0\\
 log batch step & 100 \\
 train rolling loss step & 100\\
 weight decay & 0.0\\
 learning rate & $1e-5$\\
 warmup steps & 100\\
 gradient accumulation steps & 1\\
 num train epochs & 30/60*\\
 eval per epoch & 1\\
 hard negatives & 0\\
 other negatives & 1(2,3,4,5)*\\
 val av rank hard neg & 0\\
 val av rank other neg & 10\\
 val av rank bsz & 128\\
 val av rank max qs & 10000\\
 \bottomrule
 \end{tabular}
 }
 \label{tabB01}
\end{table}

\begin{table}[ht!]
 \centering
 \caption{Hyperparameters of reader model fine-tuning.}
 \resizebox{0.89\columnwidth}{!}{
 \begin{tabular}{lr}
 \toprule
 \textbf{Hyperparameter} & \textbf{Value} \\
 \midrule
 eval step & 50/500*\\
 batch size & 32\\
 dev batch size & 32\\
 adam eps & $1e-8$\\
 adam betas & (0.9, 0.999)\\
 max grad norm & 1.0\\
 log batch step & 100 \\
 train rolling loss step & 100\\
 weight decay & 0.0\\
 learning rate & $1e-5$\\
 warmup steps & 0\\
 gradient accumulation steps & 1\\
 num train epochs & 10/30*\\
 \bottomrule
 \end{tabular}
 \label{tabB02}
 }
\end{table}

\begin{table}[h!]
\centering
\caption{Automatic evaluation of fine-tuned retrieval models using recall@1 scores when using the enhanced method of selecting negative contexts.}
\resizebox{0.95\columnwidth}{!}{
\begin{tabular}{l|cccc}
\toprule
\textbf{Methods} & \textit{\textbf{BioASQ}} & \textit{\textbf{COVID-QA}} & \textit{\textbf{cpgQA}} & \textit{\textbf{SleepQA}} \\
\hline
\textbf{{baseline}} & \textbf{25.0} & \textbf{42.5} & \textbf{66.4} & \textbf{46.8} \\
\hline
\textbf{\textit{BertScore} (1 neg)} & \color{green}31.2 & \color{red}41.6 & 66.4 & \color{green}47.2 \\
\textbf{\textit{BertScore} (2 neg)} & \color{green}28.1 & \textbf{\cellcolor{green!100}48.7} & \textbf{\cellcolor{green!100}67.3} & \color{red}45.8 \\
\textbf{\textit{BertScore} (3 neg)} & \textbf{\cellcolor{green!100}32.3} & \color{green}45.1 & \textbf{\cellcolor{green!100}67.3} &  \color{green}47.4 \\
\textbf{\textit{BertScore} (4 neg)} & \color{green}29.2 & \color{green}45.1 & \color{red}63.6 & \color{red}46.6 \\
\textbf{\textit{BertScore} (5 neg)} & \color{green}30.2 & \textbf{\cellcolor{green!100}48.7} & \color{red}61.8 & \textbf{\cellcolor{green!100}48.4} \\
\hline
\textbf{generation time} & 1.3 & 1.3 & 0.7 & 28.3 \\
\bottomrule
\end{tabular}
}
\label{tabB1}
\end{table}

\subsection{Negative Contexts}

Using the enhanced method of selecting negative contexts, we produced five different training sets for each dataset (see \Cref{tabB1}). Although generally, this method produced enhanced training sets for each dataset, it is not possible to conclude which number of negatives improved the fine-tuning process the best, as this is very much dataset-specific. The last row in \Cref{tabB1} shows the time (in hours) needed to generate all five training sets for each dataset using A100 GPU 40GB. While for most of the datasets, the generation process took around one hour, for \textit{SleepQA} it took more than one day.

\subsection{Paraphrasing}

For question paraphrasing, we used \textit{T5} and \textit{Pegasus} as they are based on Transformer architecture and utilize transfer learning, in which resource-rich sources can be efficiently adapted for resource-poor target fields, such as the domain-specific datasets \cite{yu2018modelling}. 

\begin{table}[h!]
\centering
\caption{Average similarity index of each training set for each dataset, calculated using a word vector model from spaCy for paraphrasing.}\resizebox{0.95\columnwidth}{!}{
\begin{tabular}{l|cccccc}
\toprule
\textbf{Methods} & \textbf{set 1} & \textbf{set 2} & \textbf{set 3} & \textbf{set 4} & \textbf{set 5} & \textbf{set 6}\\
\hline
\textbf{\textit{BioASQ (T5)}} & 0.997 & 0.991 & 0.979 & 0.962 & 0.927 & 0.970 \\
\textbf{\textit{BioASQ (Pegasus)}} & 0.953 & 0.932 & 0.917 & 0.886 & 0.846 & 0.903 \\
\hline
\textbf{\textit{COVID-QA (T5)}} & 0.996 & 0.987 & 0.970 & 0.949 & 0.904 & 0.959 \\
\textbf{\textit{COVID-QA (Pegasus)}} & 0.959 & 0.940 & 0.918 & 0.890 & 0.849 & 0.909 \\
\hline
\textbf{\textit{cpgQA (T5)}} & 0.995 & 0.987 & 0.973 & 0.954 & 0.920 & 0.967 \\
\textbf{\textit{cpgQA (Pegasus)}} & 0.960 & 0.946 & 0.930 & 0.910 & 0.883 & 0.925 \\
\hline
\textbf{\textit{SleepQA (T5)}} & 0.996 & 0.985 & 0.969 & 0.947 & 0.906 & 0.960 \\
\textbf{\textit{SleepQA (Pegasus)}} & 0.974 & 0.957 & 0.938 & 0.915 &  0.880 & 0.933 \\
\bottomrule
\end{tabular}
}
\label{tabB2}
\end{table}

Previous research showed that the \textit{Pegasus} method produces paraphrases that are semantically more different, while the \textit{T5} method is found to keep more of the original meaning \cite{martin2023using}. We found that the \textit{Pegasus} consistently produces the same set of paraphrased questions, regardless of the number generated. For \textit{T5}, we generated paraphrased questions up to 50 times, after which we took the first five unique paraphrases. For several questions (between 3\% for \textit{cpgQA} dataset and 12\% for \textit{COVID-QA} dataset), \textit{T5} failed to produce the required number of unique paraphrases, for which cases we added the original question to the set of five paraphrased questions. Although we used two different libraries, question paraphrasing failed to enhance training set quality for \textit{cpgQA} dataset altogether. Generating training sets took around 15 hours for \textit{SleepQA} dataset and 3 hours for other datasets on one NVIDIA TESLA P100 GPU 16GB (Kaggle).

\begin{table}[b!]
\centering
\caption{Automatic evaluation of fine-tuned retrieval models using recall@1 scores for paraphrasing. Baseline recall@1 scores for \textit{BioASQ}, \textit{COVID-QA}, \textit{cpgQA} and \textit{SleepQA} datasets are: \textbf{25.0}, \textbf{42.5}, \textbf{66.4}, and \textbf{46.8}.}
\resizebox{0.95\columnwidth}{!}{
\begin{tabular}{l|cccccc}
\toprule
\textbf{Methods} & \textbf{set 1} & \textbf{set 2} & \textbf{set 3} & \textbf{set 4} & \textbf{set 5} & \textbf{set 6}\\
\hline
\textbf{\textit{BioASQ (T5)}} & 25.0 & \textbf{\cellcolor{green!30}29.2} & \color{green}26.0 & \color{green}26.0 & \color{red}24.0 & \color{red}24.0 \\
\textbf{\textit{BioASQ (Pegasus)}} & \color{red}28.1 & \textbf{\cellcolor{green!100}31.2} & \textbf{\cellcolor{green!100}31.2} & \color{green}29.2 & \textbf{\cellcolor{green!100}31.2} & \color{green}30.2 \\
\hline
\textbf{\textit{COVID-QA (T5)}} & \color{green}49.6 & \color{green}48.7 & \color{green}44.2 & \color{green}47.8 & \color{green}46.0 & \textbf{\cellcolor{green!100}54.0} \\
\textbf{\textit{COVID-QA (Pegasus)}} &\color{green}45.1 & \color{green}44.2 & \color{green}43.4 & \color{green}43.4 & \textbf{\cellcolor{green!30}46.9} & \textbf{\color{green}46.9} \\
\hline
\textbf{\textit{cpgQA (T5)}} & \color{red}65.5 & \color{red}65.5 &\color{red}65.5 & \textbf{66.4} & \color{red}65.5 & \textbf{66.4} \\
\textbf{\textit{cpgQA (Pegasus)}} & \color{red}63.6 & \color{red}62.7 & \color{red}60.0 & \color{red}62.7 & \textbf{\cellcolor{red!50}65.5} & \color{red}69.0  \\
\hline
\textbf{\textit{SleepQA (T5)}} & \color{red}43.6 & \textbf{\cellcolor{red!20}46.6} & \color{red}42.4 & \color{red}46.4 & \color{red}44.2 & \color{red}{43.6} \\
\textbf{\textit{SleepQA (Pegasus)}} & \color{red}43.2 & \color{red}39.8 & \textbf{\cellcolor{red!50}45.0} & \color{red}39.0 & \color{red}38.0 & \color{red}41.0 \\
\bottomrule
\end{tabular}
}
\label{tabB3}
\end{table}

\subsection{Word Substitution}

\textit{Word substitution} is the process of substituting similar words (such as synonyms or words with similar embeddings) from the original data \cite{pappas2022data}. This method for enhancing the original training sets increased almost all recall@1/EM scores for all datasets for both retrieval/reader fine-tuning, except for the reader models for \textit{cpgQA} and \textit{COVID-QA} datasets. In cases where applying word substitution on the original dataset did not increase the EM scores for the reader fine-tuning, the scores stayed the same as the corresponding baselines (i.e., this method did not worsen them). Moreover, the generation of training sets took only 11 minutes for \textit{SleepQA} dataset and around two minutes for other datasets on one NVIDIA TESLA P100 GPU 16GB (Kaggle).

\begin{table}[t!]
\centering
\caption{Automatic evaluation of fine-tuned reader models using EM scores for paraphrasing. Baseline EM scores for \textit{BioASQ}, \textit{COVID-QA}, \textit{cpgQA} and \textit{SleepQA} datasets are: \textbf{5.2}, \textbf{22.1}, \textbf{50.9}, and \textbf{58.6}.}
\resizebox{1\columnwidth}{!}{
\begin{tabular}{l|cccccc}
\toprule
\textbf{Methods} & \textbf{set 1} & \textbf{set 2} & \textbf{set 3} & \textbf{set 4} & \textbf{set 5} & \textbf{set 6}\\
\hline
\textbf{\textit{BioASQ (T5)}} & \color{red}4.2 & \textbf{\cellcolor{green!30}6.2} & \color{red}4.2 & \color{red}3.1 & \textbf{\color{green}6.2} & \color{red}4.2 \\
\textbf{\textit{BioASQ (Pegasus)}} & \color{green}6.2 & \textbf{\cellcolor{green!100}7.3} & \textbf{\cellcolor{green!100}7.3} & \color{green}6.2 & \color{green}6.2 & \color{green}6.2 \\
\hline
\textbf{\textit{COVID-QA (T5)} }& \color{red}21.2 & \color{red}19.5 & \color{red}20.4 & \textbf{\cellcolor{green!100}23.9} & \color{red}20.4 & \color{red}19.5 \\
\textbf{\textit{COVID-QA (Pegasus)}} & 22.1 & \color{red}18.6 & \color{red}18.6 & \color{red}20.4 & \textbf{\cellcolor{green!30}23.0} & \color{red}19.5 \\
\hline
\textbf{\textit{cpgQA (T5)}} & \textbf{50.9} & \color{red}49.1 & \color{red}48.2 & \textbf{50.9} & \color{red}48.2 & \color{red}50.0 \\
\textbf{\textit{cpgQA (Pegasus)}} & \color{red}46.4 & \color{red}46.4 & \color{red}47.3 & \color{red}44.5 & \color{red}46.4 & \textbf{\cellcolor{green!100}49.1} \\
\hline
\textbf{\textit{SleepQA (T5)}} & \color{red}57.4 & \color{red}57.6 & \color{red}58.2 & \color{red}58.4 & \textbf{\cellcolor{green!30}58.8} & \color{red}{58.2} \\
\textbf{\textit{SleepQA (Pegasus)}} & \color{red}58.2 & \color{red}57.8 & \color{red}58.0 & \color{red}58.2 & \color{red}57.2 & \textbf{\cellcolor{green!100}59.0} \\
\bottomrule
\end{tabular}
}
\label{tabB4}
\end{table}

\begin{table}[h!]
\centering
\caption{Average similarity index of each training set for each dataset, calculated using a word vector model from spaCy for word substitution.}\resizebox{1\columnwidth}{!}{
\begin{tabular}{l|cccccc}
\toprule
\textbf{Datasets} & \textbf{set 1} & \textbf{set 2} & \textbf{set 3} & \textbf{set 4} & \textbf{set 5} & \textbf{set 6}\\
\hline
\textbf{\textit{BioASQ}} & 0.999 & 0.998 & 0.997 & 0.996 & 0.994 & 0.997 \\
\textbf{\textit{COVID-QA}} & 0.997 & 0.996 & 0.995 & 0.993 & 0.988 & 0.993 \\
\textbf{\textit{cpgQA}} & 0.998 & 0.997 & 0.996 & 0.994 & 0.989 & 0.995 \\
\textbf{\textit{SleepQA}} & 0.996 & 0.993 & 0.992 & 0.990 & 0.986 & 0.991 \\
\bottomrule
\end{tabular}
}
\label{tabB5}
\end{table}

\begin{table}[h!]
\centering
\caption{Automatic evaluation of fine-tuned retrieval models using recall@1 for word substitution. Baseline recall@1 scores for \textit{BioASQ}, \textit{COVID-QA}, \textit{cpgQA} and \textit{SleepQA} datasets are: \textbf{25.0}, \textbf{42.5}, \textbf{66.4}, and \textbf{46.8}.}
\resizebox{1\columnwidth}{!}{
\begin{tabular}{l|cccccc}
\toprule
\textbf{Datasets} & \textbf{set 1} & \textbf{set 2} & \textbf{set 3} & \textbf{set 4} & \textbf{set 5} & \textbf{set 6}\\
\hline
\textbf{\textit{BioASQ}} & \color{green}28.1 & \color{red}24.0 & \color{green}28.1 & \color{green}27.1 & \textbf{\cellcolor{green!100}30.2} & \color{red}21.9 \\
\textbf{\textit{COVID-QA}} & \color{green}49.6 & \color{green}49.6 & \textbf{\cellcolor{green!100}50.4} & \color{green}46.9 & \color{green}48.7 & \color{green}48.7 \\
\textbf{\textit{cpgQA}} & \color{red}63.6 & \color{green}68.2 & \color{green}67.3 & \textbf{\cellcolor{green!100}69.1} & \color{green}67.3 & 66.4 \\
\textbf{\textit{SleepQA}} & \color{red}45.8 & \textbf{\cellcolor{green!100}48.4} & \color{red}46.4 & 46.8 & \color{red}43.0 & \color{red}46.0 \\
\bottomrule
\end{tabular}
}
\label{tabB6}
\end{table}

\begin{table}[b!]
\centering
\caption{Automatic evaluation of fine-tuned reader models using EM scores for word substitution. Baseline EM for \textit{BioASQ}, \textit{COVID-QA}, \textit{cpgQA} and \textit{SleepQA} datasets are: \textbf{5.2}, \textbf{22.1}, \textbf{50.9}, and \textbf{58.6}.}
\resizebox{\columnwidth}{!}{
\begin{tabular}{l|cccccc}
\toprule
\textbf{Datasets} & \textbf{set 1} & \textbf{set 2} & \textbf{set 3} & \textbf{set 4} & \textbf{set 5} & \textbf{set 6}\\
\hline
\textbf{\textit{BioASQ}} & 5.2 & \textbf{\cellcolor{green!100}6.2} & 5.2 & 5.2 & \textbf{\cellcolor{green!100}6.2} & \textbf{\cellcolor{green!100}6.2} \\
\textbf{\textit{COVID-QA}} & \color{red}21.2 & \color{red}21.2 & \color{red}21.2 & \textbf{22.1} & \color{red}21.2 & \color{red}19.5 \\
\textbf{\textit{cpgQA}} & \color{red}50.0 & \color{red}50.0 & \textbf{50.9} & \color{red}50.0 & \textbf{50.9} & \textbf{50.9} \\
\textbf{\textit{SleepQA}} & \color{red}57.8 & 58.6 & \color{green}58.8 & \textbf{\cellcolor{green!100}59.4} & \color{red}58.0 & \color{red}58.0 \\
\bottomrule
\end{tabular}
}
\label{tabB7}
\end{table}

\subsection{Back Translation}

The main idea behind back translation method is to use machine translation from a source to a pivot language and back, obtaining paraphrases. In total, we generated 25 different training sets for Spanish (es), French (fr), German (de), Russian (ru), Chinese (zh), Arabic (ar), Dutch (nl), Finnish (fi), Hungarian (hu), Multiple Languages (mul), Ukrainian (uk), Hindi (hi), Danish (da), Czech (cs), Romance Languages (roa), Bulgarian (bg), Catalan (ca), Afrikaans (af), Estonian (et), Turkic Languages (trk), Slavik Languages (sla), Indonesian (id), Slovak (sk), Tagalog (tl), and Kinyarwanda (rw) pivot languages. Back translation has been used as a data augmentation method for several different NLP tasks \cite{feng2021survey, shorten2021text}. Generally, it produced the best results for \textit{BioASQ} dataset. The generation of training sets took 10 hours for \textit{SleepQA} dataset and around two hours for other datasets on one NVIDIA TESLA P100 GPU 16GB (Kaggle). Results are in \Cref{tabB8} and \Cref{tabB9}.

\clearpage

\begin{table}[t!]
\centering
\caption{Automatic evaluation of fine-tuned retrieval models using recall@1 for back translation. Baseline recall@1 scores for \textit{BioASQ}, \textit{COVID-QA}, \textit{cpgQA} and \textit{SleepQA} datasets are: \textbf{25.0}, \textbf{42.5}, \textbf{66.4}, and \textbf{46.8}.}
\resizebox{0.92\columnwidth}{!}{
\begin{tabular}{l|rrrr}
\toprule
\textbf{Methods} & \textit{\textbf{BioASQ}} & \textit{\textbf{COVID-QA}} & \textit{\textbf{cpgQA}} & \textit{\textbf{SleepQA}} \\
\hline
\textbf{en-es-en} & \color{green}31.2 & \color{green}48.7 & \color{red}62.7 & \color{red}45.4 \\
\textbf{en-fr-en} & \color{green}29.2 & \color{green}47.8 & \color{red}60.9 & \color{red}44.8 \\
\textbf{en-de-en} & \color{green}27.1 & \color{green}45.1 & \color{red}61.8 & \color{red}41.6 \\
\textbf{en-ru-en} & \color{green}31.2 & \color{red}40.7 & \color{red}54.5 & \color{red}39.8 \\
\textbf{en-zh-en} & \color{green}30.2 & \color{green}46.9 & \color{red}61.8 & \color{red}42.2 \\
\textbf{en-ar-en} & \color{green}30.2 & \color{green}49.6 & \color{red}56.4 & \color{red}41.2 \\
\textbf{en-nl-en} & \color{green}31.2 & \color{red}40.7 & \color{red}64.5 & \color{red}44.8 \\
\textbf{en-fi-en} & \color{green}27.1 & \color{green}48.7 & \color{red}61.8 & \color{red}40.6 \\
\textbf{en-hu-en} & \color{green}29.2 & \textbf{\cellcolor{green!100}49.6} & \textbf{66.4} & \color{red}41.6 \\
\textbf{en-mul-en} & 25.0 & \color{green}43.4 & \color{red}57.3 & \color{red}39.4 \\
\textbf{en-uk-en} & \color{green}28.1 & \color{green}45.1 & \color{red}64.5 & \color{red}40.8 \\
\textbf{en-hi-en} & \color{green}27.1 & \color{green}44.2 & \color{red}59.1 & \color{red}38.4 \\
\textbf{en-da-en} & \color{green}29.2 & \color{green}44.2 & \color{red}60.0 & \color{red}43.8 \\
\textbf{en-cs-en} & \color{green}27.1 & \color{green}43.4 & \color{red}63.6 & \textbf{\cellcolor{red!50}45.8} \\
\textbf{en-roa-en} & \color{green}29.2 & \color{green}47.8 & \color{red}60.9 & \color{red}42.0 \\
\textbf{en-bg-en} & \color{green}29.2 & \color{green}43.4 & \color{red}58.2 & \color{red}40.0 \\
\textbf{en-ca-en} & \textbf{\cellcolor{green!100}33.3} & \color{red}41.6 & \color{red}60.0 & \color{red}41.2 \\
\textbf{en-af-en} & \color{green}30.2 & \color{green}46.9 & \color{red}61.8 & \color{red}37.2 \\
\textbf{en-et-en} & \color{green}29.2 & \color{green}46.0 & \color{red}58.2 & \color{red}40.2 \\
\textbf{en-trk-en} & \color{red}18.8 & \color{red}23.9 & \color{red}35.5 & \color{red}19.6 \\
\textbf{en-sla-en} & 25.0 & \color{green}45.1 & \color{red}63.6 & \color{red}43.6 \\
\textbf{en-id-en} & \color{green}30.2 & \color{green}47.8 & \color{red}63.6 & \color{red}40.4 \\
\textbf{en-sk-en} & \color{green}30.2 & \color{green}48.7 & \color{red}57.3 & \color{red}44.2 \\
\textbf{en-tl-en} & \color{green}30.2 & \color{red}41.6 & \color{red}64.5 & \color{red}40.8 \\
\textbf{en-rw-en} & \color{green}28.1 & \color{red}29.2 & \color{red}50.0 & \color{red}34.4 \\
\bottomrule
\end{tabular}
}
\label{tabB8}
\end{table}

\begin{table}[t!]
\centering
\caption{Automatic evaluation of fine-tuned reader models using EM scores for back translation. Baseline EM scores for \textit{BioASQ}, \textit{COVID-QA}, \textit{cpgQA} and \textit{SleepQA} datasets are: \textbf{5.2}, \textbf{22.1}, \textbf{50.9}, and \textbf{58.6}.}
\resizebox{0.92\columnwidth}{!}{
\begin{tabular}{l|cccc}
\toprule
\textbf{Methods} & \textit{\textbf{BioASQ}} & \textit{\textbf{COVID-QA}} & \textit{\textbf{cpgQA}} & \textit{\textbf{SleepQA}} \\
\hline
\textbf{en-es-en} & \color{red}4.2 & \color{red}21.2 & \color{red}40.0 & \color{red}58.2 \\
\textbf{en-fr-en} & \color{green}6.2 & \color{red}20.4 & \color{red}45.5 & \color{red}58.4 \\
\textbf{en-de-en} & \textbf{\cellcolor{green!100}7.3} & \color{red}21.2 & \textbf{\cellcolor{red!50}46.4} & \color{red}57.4 \\
\textbf{en-ru-en} & \color{red}3.1 & \color{red}18.6 & \color{red}45.5 & \color{red}58.4 \\
\textbf{en-zh-en} & \color{green}6.2 & \color{red}21.2 & \color{red}43.6 & \color{green}58.8 \\
\textbf{en-ar-en} & 5.2 & \textbf{\cellcolor{green!100}23.0} & \color{red}44.5 & \color{red}58.2 \\
\textbf{en-nl-en} & \textbf{\cellcolor{green!100}7.3} & \color{red}21.2 & \color{red}45.5 & \color{red}57.6 \\
\textbf{en-fi-en} & \color{green}6.2 & \color{red}20.4 & \color{red}44.5 & \color{red}58.0 \\
\textbf{en-hu-en} & \color{green}6.2 & \color{red}19.5 & \color{red}43.6 & \color{red}58.2 \\
\textbf{en-mul-en} & \color{red}3.1 & \color{red}19.5 & \color{red}43.6 & \color{red}57.0 \\
\textbf{en-uk-en} & \color{green}6.2 & \color{red}18.6 & \color{red}40.9 & \textbf{\cellcolor{green!100}59.4} \\
\textbf{en-hi-en} & 5.2 & \color{red}20.4 & \color{red}40.9 & \color{red}57.4 \\
\textbf{en-da-en} & \color{green}6.2 & \textbf{\cellcolor{green!100}23.0} & \color{red}43.6 & \cellcolor{green!100}\textbf{59.4} \\
\textbf{en-cs-en} & \color{red}4.2 & \color{red}19.5 & \color{red}43.6 & \color{red}58.0 \\
\textbf{en-roa-en} & \color{green}6.2 & \color{red}18.6 & \color{red}43.6 & \color{red}57.6 \\
\textbf{en-bg-en} & \color{green}6.2 & \color{red}21.2 & \color{red}43.6 & \color{green}59.2 \\
\textbf{en-ca-en} & 5.2 & \color{red}18.6 & \color{red}43.6 & \color{red}58.2 \\
\textbf{en-af-en} & \textbf{\cellcolor{green!100}7.3} & \color{red}20.4 & \color{red}44.5 & \color{green}59.0 \\
\textbf{en-et-en} & \color{green}6.2 & \color{red}20.4 & \color{red}43.6 & \color{red}58.0 \\
\textbf{en-trk-en} & \color{red}4.2 & \color{red}15.9 & \color{red}39.1 & \color{red}56.4 \\
\textbf{en-sla-en} & \color{green}6.2 & \color{red}18.6 & \color{red}44.5 & \color{red}57.6 \\
\textbf{en-id-en} & \color{red}3.1 & \color{red}17.7 & \color{red}44.5 & \color{red}57.2 \\
\textbf{en-sk-en} & 5.2 & \color{red}21.2 & \color{red}44.5 & 58.6 \\
\textbf{en-tl-en} & \color{red}4.2 & 22.1 & \color{red}46.4 & \color{red}58.4 \\
\textbf{en-rw-en} & 5.2 & \color{red}17.7 & \color{red}40.0 & \color{red}56.2 \\
\bottomrule
\end{tabular}
}
\label{tabB9}
\end{table}

\subsection{Mean and Standard Deviation}
\label{subsec:b6}

\Cref{tabB10} shows the mean and standard deviation for different data quality enhancement methods for retrieval fine-tuning. \Cref{tabB11} shows the mean and standard deviation for different data quality enhancement methods for reader fine-tuning. 

\begin{table}[H]
\centering
\caption{Mean and standard deviation of different data quality enhancement methods for retrieval fine-tuning.}
\resizebox{\columnwidth}{!}{
\begin{tabular}{l|cccc}
\toprule
\textbf{Methods} & \textit{\textbf{BioASQ}} & \textit{\textbf{COVID-QA}} & \textit{\textbf{cpgQA}} & \textit{\textbf{SleepQA}} \\
\hline
\textbf{\textit{negatives}} & $30.2 \pm 1.7 $ & $45.8 \pm 3.0$  & $66.0 \pm 2.4$ & $47.1 \pm 1.0$ \\
\hline
\textbf{\textit{paraphrasing (T5)}} & $25.7 \pm 1.9$ & $48.4 \pm 3.4$ & $65.8 \pm 0.5$ & $44.5 \pm 1.7$ \\
\textbf{\textit{paraphrasing (Pegasus)}} & $30.2 \pm 1.3$ & $45.0 \pm 1.6$ & $64.0 \pm 3.1$ & $41.0 \pm 2.7$ \\
\hline
\textbf{\textit{substitution}} & $26.6 \pm 3.1 $ & $49.0 \pm 1.2$ & $67.0 \pm 1.9$ & $46.1 \pm 1.8$ \\
\hline
\textit{\textbf{translation}} & $28.7 \pm 2.8$ & $44.0 \pm 6.0$ & $ 59.6 \pm 6.2 $ & $40.6 \pm 5.1$ \\
\bottomrule
\end{tabular}
}
\label{tabB10}
\end{table}

\begin{table}[H]
\centering
\caption{Mean and standard deviation of different data quality enhancement methods for reader fine-tuning.}
\resizebox{\columnwidth}{!}{
\begin{tabular}{l|cccc}
\toprule
\textbf{Methods} & \textit{\textbf{BioASQ}} & \textit{\textbf{COVID-QA}} & \textit{\textbf{cpgQA}} & \textit{\textbf{SleepQA}} \\
\hline
\textbf{\textit{paraphrasing (T5)}} & $ 4.7 \pm 1.3 $ & $ 20.8 \pm 1.6 $ & $ 49.6 \pm 1.2 $ & $ 58.1 \pm 0.6 $ \\
\textbf{\textit{paraphrasing (Pegasus)}} & $ 6.6 \pm 0.5 $ & $ 20.4 \pm 1.8 $ & $ 46.7 \pm 1.5 $ & $ 58.1 \pm 0.6 $ \\
\hline
\textbf{\textit{substitution}} & $ 5.7 \pm 0.5 $ & $ 21.1 \pm 0.8 $ & $ 50.5 \pm 0.5 $ & $ 58.4 \pm 0.6 $ \\
\hline
\textit{\textbf{translation}} & $ 5.4 \pm 1.3 $ & $ 20.0 \pm 1.7 $ & $ 43.6 \pm 2.0 $ & $ 58.0 \pm 0.8 $ \\
\bottomrule
\end{tabular}
}
\label{tabB11}
\end{table}

\subsection{Similarity Between Enhancement Methods}
\label{subsec:b7}

In the following tables, we show the average similarity computed with ROUGE-1 metric between questions generated through each of the enhancement techniques, over all four datasets \{\emph{BioASQ},\emph{CovidQA},\emph{cpgQA},\emph{SleepQA}\}, with Retrieval (first four tables) then Reader (next four).

\begin{table}[H]
\centering
\caption{\textbf{Average ROUGE-1 score} between pairs of questions from different enhancement methods on \textbf{BioASQ retrieval datasets}. \textbf{Base.} stands for baseline, \textbf{Para/PG} for paraphrasing with PEGASUS, \textbf{Para/T5} for paraphrasing with T5, \textbf{Subst.} for substitution and \textbf{Transl.} for translation.
}
\renewcommand{\arraystretch}{0}
\setlength{\fboxsep}{3mm} 
\setlength{\tabcolsep}{0pt}
\begin{center}
\resizebox{0.99\columnwidth}{!}{
\begin{tabular}{c*{5}{R}}
  & \textbf{Base.} & \textbf{Para/PG} & \textbf{Para/T5} & \textbf{Subst.} & \textbf{Transl.} \EndTableHeader\\
  \noalign{\vskip 1mm} 
  \textbf{Base.}~~   & 100.0 \\
  \textbf{Para/PG}~~ & 80.44 & 100.0 \\
  \textbf{Para/T5}~~ & 91.49 & 76.73 & 100.0 \\
  \textbf{Subst.}~~  & 95.11 & 76.25 & 86.98 & 100.0 \\
  \textbf{Transl.}~~ & 57.68 & 51.10 & 56.98 & 55.01 & 100.0 \\
\end{tabular}
}
\label{tabB12}
\end{center}
\end{table}

\begin{table}[H]
\centering
\caption{\textbf{Average ROUGE-1 score} between pairs of questions from different enhancement methods on \textbf{CovidQA retrieval datasets}. 
}
\renewcommand{\arraystretch}{0}
\setlength{\fboxsep}{3mm} 
\setlength{\tabcolsep}{0pt}
\begin{center}
\resizebox{0.99\columnwidth}{!}{
\begin{tabular}{c*{5}{R}}
  & \textbf{Base.} & \textbf{Para/PG} & \textbf{Para/T5} & \textbf{Subst.} & \textbf{Transl.} \EndTableHeader\\
  \noalign{\vskip 1mm} 
  \textbf{Base.}~~   & 100.0 \\
  \textbf{Para/PG}~~ & 74.63 & 100.0 \\
  \textbf{Para/T5}~~ & 83.33 & 66.06 & 100.0 \\
  \textbf{Subst.}~~  & 95.33 & 70.89 & 79.73 & 100.0 \\
  \textbf{Transl.}~~ & 76.44 & 62.60 & 69.50 & 72.97 & 100.0 \\
\end{tabular}
}
\label{tabB13}
\end{center}
\end{table}

\begin{table}[H]
\centering
\caption{\textbf{Average ROUGE-1 score} between pairs of questions from different enhancement methods on \textbf{cpgQA retrieval datasets}. 
}
\renewcommand{\arraystretch}{0}
\setlength{\fboxsep}{3mm} 
\setlength{\tabcolsep}{0pt}
\begin{center}
\resizebox{0.99\columnwidth}{!}{
\begin{tabular}{c*{5}{R}}
  & \textbf{Base.} & \textbf{Para/PG} & \textbf{Para/T5} & \textbf{Subst.} & \textbf{Transl.} \EndTableHeader\\
  \noalign{\vskip 1mm} 
  \textbf{Base.}~~   & 100.0 \\
  \textbf{Para/PG}~~ & 71.08 & 100.0 \\
  \textbf{Para/T5}~~ & 80.96 & 62.79 & 100.0 \\
  \textbf{Subst.}~~  & 94.62 & 67.01 & 76.93 & 100.0 \\
  \textbf{Transl.}~~ & 71.06 & 58.85 & 64.82 & 67.36 & 100.0 \\
\end{tabular}
}
\label{tabB14}
\end{center}
\end{table}

\begin{table}[H]
\centering
\caption{\textbf{Average ROUGE-1 score} between pairs of questions from different enhancement methods on \textbf{SleepQA retrieval datasets}. 
}
\renewcommand{\arraystretch}{0}
\setlength{\fboxsep}{3mm} 
\setlength{\tabcolsep}{0pt}
\begin{center}
\resizebox{0.99\columnwidth}{!}{
\begin{tabular}{c*{5}{R}}
  & \textbf{Base.} & \textbf{Para/PG} & \textbf{Para/T5} & \textbf{Subst.} & \textbf{Transl.} \EndTableHeader\\
  \noalign{\vskip 1mm} 
  \textbf{Base.}~~   & 100.0 \\
  \textbf{Para/PG}~~ & 77.43 & 100.0 \\
  \textbf{Para/T5}~~ & 86.30 & 70.98 & 100.0 \\
  \textbf{Subst.}~~  & 92.95 & 71.09 & 79.79 & 100.0 \\
  \textbf{Transl.}~~ & 79.05 & 65.98 & 73.53 & 73.20 & 100.0 \\
\end{tabular}
}
\label{tabB15}
\end{center}
\end{table}

\begin{table}[H]
\centering
\caption{\textbf{Average ROUGE-1 score} between pairs of questions from different enhancement methods on \textbf{BioASQ reader datasets}. 
}
\renewcommand{\arraystretch}{0}
\setlength{\fboxsep}{3mm} 
\setlength{\tabcolsep}{0pt}
\begin{center}
\resizebox{0.99\columnwidth}{!}{
\begin{tabular}{c*{5}{R}}
  & \textbf{Base.} & \textbf{Para/PG} & \textbf{Para/T5} & \textbf{Subst.} & \textbf{Transl.} \EndTableHeader\\
  \noalign{\vskip 1mm} 
  \textbf{Base.}~~   & 100.0 \\
  \textbf{Para/PG}~~ & 80.44 & 100.0 \\
  \textbf{Para/T5}~~ & 91.49 & 76.73 & 100.0 \\
  \textbf{Subst.}~~  & 97.71 & 78.51 & 89.46 & 100.0 \\
  \textbf{Transl.}~~ & 86.72 & 72.32 & 82.66 & 84.86 & 100.0 \\
\end{tabular}
}
\label{tabB16}
\end{center}
\end{table}

\begin{table}[H]
\centering
\caption{\textbf{Average ROUGE-1 score} between pairs of questions from different enhancement methods on \textbf{CovidQA reader datasets}. 
}
\renewcommand{\arraystretch}{0}
\setlength{\fboxsep}{3mm} 
\setlength{\tabcolsep}{0pt}
\begin{center}
\resizebox{0.99\columnwidth}{!}{
\begin{tabular}{c*{5}{R}}
  & \textbf{Base.} & \textbf{Para/PG} & \textbf{Para/T5} & \textbf{Subst.} & \textbf{Transl.} \EndTableHeader\\
  \noalign{\vskip 1mm} 
  \textbf{Base.}~~   & 100.0 \\
  \textbf{Para/PG}~~ & 72.11 & 100.0 \\
  \textbf{Para/T5}~~ & 83.33 & 64.65 & 100.0 \\
  \textbf{Subst.}~~  & 93.84 & 67.50 & 78.59 & 100.0 \\
  \textbf{Transl.}~~ & 66.01 & 54.99 & 60.87 & 61.55 & 100.0 \\
\end{tabular}
}
\label{tabB17}
\end{center}
\end{table}

\begin{table}[H]
\centering
\caption{\textbf{Average ROUGE-1 score} between pairs of questions from different enhancement methods on \textbf{cpgQA reader datasets}. 
}
\renewcommand{\arraystretch}{0}
\setlength{\fboxsep}{3mm} 
\setlength{\tabcolsep}{0pt}
\begin{center}
\resizebox{0.99\columnwidth}{!}{
\begin{tabular}{c*{5}{R}}
  & \textbf{Base.} & \textbf{Para/PG} & \textbf{Para/T5} & \textbf{Subst.} & \textbf{Transl.} \EndTableHeader\\
  \noalign{\vskip 1mm} 
  \textbf{Base.}~~   & 100.0 \\
  \textbf{Para/PG}~~ & 79.46 & 100.0 \\
  \textbf{Para/T5}~~ & 86.09 & 72.62 & 100.0 \\
  \textbf{Subst.}~~  & 95.58 & 75.34 & 82.15 & 100.0 \\
  \textbf{Transl.}~~ & 80.67 & 68.91 & 75.40 & 77.41 & 100.0 \\
\end{tabular}
}
\label{tabB18}
\end{center}
\end{table}

\begin{table}[H]
\centering
\caption{\textbf{Average ROUGE-1 score} between pairs of questions from different enhancement methods on \textbf{SleepQA reader datasets}. 
}
\renewcommand{\arraystretch}{0}
\setlength{\fboxsep}{3mm} 
\setlength{\tabcolsep}{0pt}
\begin{center}
\resizebox{0.99\columnwidth}{!}{
\begin{tabular}{c*{5}{R}}
  & \textbf{Base.} & \textbf{Para/PG} & \textbf{Para/T5} & \textbf{Subst.} & \textbf{Transl.} \EndTableHeader\\
  \noalign{\vskip 1mm} 
  \textbf{Base.}~~   & 100.0 \\
  \textbf{Para/PG}~~ & 68.15 & 100.0 \\
  \textbf{Para/T5}~~ & 85.57 & 62.76 & 100.0 \\
  \textbf{Subst.}~~  & 90.92 & 61.59 & 77.38 & 100.0 \\
  \textbf{Transl.}~~ & 63.06 & 51.40 & 59.00 & 57.15 & 100.0 \\
\end{tabular}
}
\label{tabB19}
\end{center}
\end{table}

\end{document}